\documentclass{article}

    \PassOptionsToPackage{numbers, compress}{natbib}

 \usepackage[main, final]{neurips_2026}

\usepackage[table,xcdraw]{xcolor}
\usepackage{multirow}
\usepackage{microtype}
\usepackage{graphicx}
\usepackage{subcaption}
\usepackage{booktabs} 
\usepackage{enumitem}

\usepackage{booktabs}
\usepackage{multirow}
\usepackage{tabularx}
\usepackage{array}
\usepackage{xcolor}
\usepackage{colortbl}
\usepackage{algorithm}
\usepackage{algpseudocode}
\usepackage{wrapfig}
\usepackage{amsmath}
\usepackage{amssymb}
\usepackage{mathtools}
\usepackage{amsthm}
\usepackage{hyperref}
\usepackage{tikz}

\newcommand{\circnum}[1]{%
\tikz[baseline=(char.base)]{
\node[shape=circle, fill=black, text=white, inner sep=0.7pt] (char) {\small #1};
}}
\usepackage[capitalize,noabbrev]{cleveref}

\DeclareMathOperator*{\argmin}{arg\,min}
\newcommand{\eg}{e.g.,~}

\newcommand{\ie}{i.e.}

\theoremstyle{plain}

\theoremstyle{definition}

\theoremstyle{remark}

\usepackage[textsize=tiny]{todonotes}
\usepackage[utf8]{inputenc} 
\usepackage[T1]{fontenc}    
\usepackage{hyperref}       
\usepackage{url}            
\usepackage{booktabs}       
\usepackage{amsfonts}       
\usepackage{nicefrac}       
\usepackage{microtype}      
\usepackage{xcolor}         

\title{MoBiQuant: Mixture-of-Bits Quantization for Token-Adaptive Any-Precision LLM}

\author{
\normalfont
  \textbf{Dongwei Wang}$^{1}$\thanks{Equal contribution.}\quad
  \textbf{Jinhee Kim}$^{2}$\footnotemark[1]\quad
  \textbf{Seokho Han}$^{3}$\footnotemark[1] \\
  \textbf{Denis Gudovskiy}$^{4}$\quad 
  \textbf{Yohei Nakata}$^{5}$\quad
  \textbf{Tomoyuki Okuno}$^{5}$\quad
  \textbf{KhayTze Peong}$^{5}$\\
  \textbf{Kang Eun Jeon}$^{6}$\thanks{Corresponding author.}\quad
  \textbf{Jong Hwan Ko}$^{3}$\footnotemark[2]\quad
  \textbf{Yiran Chen}$^{2}$\footnotemark[2]\quad
  \textbf{Huanrui Yang}$^{1}$\footnotemark[2]
  \\
  \\
  $^{1}$University of Arizona \quad
  $^{2}$Duke University \quad
  $^{3}$Sungkyunkwan University
  \\
  $^{4}$Panasonic AI Lab \quad
  $^{5}$Panasonic Holdings Corporation \quad
  $^{6}$DGIST
  \\
  \texttt{\{dongweiw,huanruiyang\}@arizona.edu} \\
  \texttt{jinhee.kim@duke.edu} \quad \texttt{beppa2396@skku.edu}
}
\begin{document}

\maketitle

\begin{abstract}
Dynamic runtime latency and memory constraints necessitate flexible large language model (LLM) deployment,  where an LLM can be inferred with various quantization precisions based on available computational resources. 
Recent work on such any-precision quantization either relies on hardware-inefficient vector quantization or induces additional scaling factors when switching between bit-widths. 
Meanwhile, existing post-training quantization (PTQ) methods calibrated for a fixed low precision show poor generalizability under runtime precision change. 
In this work, we attribute the source of poor generalization across bit-widths to a precision-dependent \textit{outlier migration} phenomenon where the distribution of PTQ-sensitive tokens changes across precisions. 
Motivated by this observation, we propose \texttt{MoBiQuant}, a novel any-precision Mixture-of-Bits quantization framework that adjusts weight precision for flexible LLM inference based on token sensitivity.
Specifically, we propose a many-in-one recursive residual quantization that can iteratively reconstruct higher-precision weights at runtime and mitigates \textit{outlier migration} with a token-aware router to dynamically select the optimal inference precision of each token.
Extensive experiments show that \texttt{MoBiQuant} matches or surpasses frontier single-precision PTQ while exhibiting strong elasticity, achieving significant memory savings and throughput gains of up to $1.34\times$ over state-of-the-art any-precision methods. Our code is released at this
\href{https://github.com/SimWangArizona/MoBiQuant}{\texttt{link}}.
\end{abstract}

\section{Introduction}
\label{sec:Introduction}
Recent deployments of large language models (LLMs)~\citep{touvron2023llama, openai2023gpt4, gemini15, deepseekv3, singhal2025llamanemotron} typically use low-precision quantization to meet  latency and memory constraints. 
Although ongoing work on post-training quantization (PTQ)~\citep{frantar2023optq, shao2023omniquant} constantly improves LLM performance at predetermined fixed precisions, the tradeoff between precision and performance remains critical. 

The impact of such a tradeoff is more significant for edge devices where available computational resources can vary dynamically at runtime. Under sufficient resources, the device can run the model at higher precision for better performance, while under resource contention from other applications, it may need to reduce precision to satisfy latency and memory constraints. 
Storing LLM weights for each precision is impractical, and, hence, recent research explores any-precision quantization where a single model can support multiple precisions~\cite{park2024any, park2025anybcq, nair2025matryoshka}.
However, existing approaches remain limited in practice, as they either rely on hardware-unfriendly vector quantization (VQ)~\cite{park2024any} or introduce additional calibration parameters that lead to training and inference overhead~\cite{park2025anybcq}. Moreover, they often lack flexibility for runtime adaptation, requiring offline repacking and kernel redeployment when switching precisions~\citep{nair2025matryoshka}. 

In this work, we overcome these limitations and propose a more advanced any-precision framework with three key properties: (1) it supports hardware-friendly scalar quantization, (2) it introduces minimal overhead compared to a single set of PTQ calibration parameters, and (3) it enables efficient runtime precision switching without offline repacking or kernel relaunching.
To achieve this, we start by investigating the root cause of why existing single-precision scalar PTQ methods fail to generalize across bit-widths with fixed calibration parameters.

We identify a critical phenomenon of precision-dependent \textit{``outlier migration''} in LLMs. Although it is well-established that outliers in activations govern quantization errors \citep{dettmers2022gptint}, we find that the specific subset of tokens that are responsible for high quantization errors are not static for each precision. 
Consequently, static PTQ methods that calibrate quantization parameters (\eg scaling factors or clipping ranges) for a single precision, often fail to generalize. They inevitably overfit to an improper set of tokens when the bit-width changes, leading to suboptimal performance in any-precision inference setting.

Motivated by this, we propose \texttt{MoBiQuant}, an any-precision quantization framework that mitigates ``\textit{outlier migration}'' with a single set of calibration parameters by incorporating token-adaptive precision adjustment during calibration. \texttt{MoBiQuant} learns to dynamically assign proper inference precision to tokens on the fly based on their sensitivity while maintaining overall inference cost within an average budget. Our approach consists of two core components. The first one is \texttt{MoBiSlice}, a ``many-in-one" recursive residual quantization that decomposes weights into a low-precision base and successive residual bit slices. This allows a single model to support multiple precisions (\eg 2, 4, 6-bit) via activating fixed-precision bit slices (\eg 2-bit slice) at runtime, thereby supporting any-precision inference with a single type of low-precision bit slice GEMM kernel without the need for repacking and redeployment.
The second component is \texttt{MoBiRoute}, a lightweight learnable router that implements token-aware inference precision selection. Acting as a gate, the \texttt{MoBiRoute} dynamically activates the optimal number of \texttt{MoBiSlice} residual components for each token based on learned token sensitivity behavior and current resource conditions, 
thereby preventing the calibration parameters from overfitting to the outlier distribution of a single precision. Runtime precision switching is achieved by adjusting the routing threshold, without introducing extra scaling factors and training overhead.

In summary, our paper makes the following contributions:
\begin{itemize}
\item The \texttt{MoBiSlice} recursive quantization explicitly reconstructs higher precisions using low-precision bit slices. It enables a single model to flexibly realize multiple precision levels via hierarchical reconstruction, without requiring offline repacking and redeployment.
\item The \texttt{MoBiRoute} router mitigates the bit-dependent \textit{outlier migration} phenomenon by dynamically selecting the number of bit slices for each token. It also enables efficient runtime precision switching by a threshold adjustment, eliminating the need to store additional scaling factors and extra training overhead.
\item Extensive experiments demonstrate that \texttt{MoBiQuant} generalizes effectively across bit-widths, matching or surpassing the performance of state-of-the-art single-precision PTQ methods. Our tailored kernel design achieves $1.34\times$ higher throughput than state-of-the-art any-precision methods~\citep{park2024any, park2025anybcq}. Compared to deploying multiple models, \texttt{MoBiQuant} reduces the memory footprint by $3.5\times$.
\end{itemize}
\section{Related Work}\label{sec:Related}

\textbf{Single-precision PTQ.} Most existing methods focus on statically-defined quantization schemes, where LLM performance is particularly optimized for the selected bit-width configuration. 
Representative works include: 
GPTQ \citep{frantar2022gptq} with second-order information to minimize quantization error, 
OmniQuant \citep{shao2023omniquant} with learnable weight clipping to achieve superior results, 
SmoothQuant \citep{xiao2023smoothquant} and SpinQuant \citep{liu2024spinquant} with special linear transformations to mitigate peaking outliers, 
and AWQ \citep{lin2024awq} for selectively quantizing salient weights. 
Recent advancements like the QuIP family of methods \citep{NEURIPS2023_0df38cd1, tseng2024quip} achieve extreme compression ratios using VQ, yet these methods suffer from limited practicality, as their VQ scheme is not well supported by modern GEMM kernels for runtime acceleration.
While these methods achieve strong performance at a given precision, they still exhibit an inherent trade-off between accuracy and efficiency across bit-widths. Moreover, they lack flexibility, as switching between precisions typically requires offline recalibration, making them unsuitable for dynamic edge scenarios with changing resource constraints.

\textbf{Any-precision PTQ.} To address the inflexibility of static schemes, recent works propose any-precision quantization, enabling a single model to support multiple bit-widths for dynamic system requirements. 
AnyPrecisionLLM \citep{park2024any} incrementally upscales a low-bit model to higher precisions, AnyBCQ \citep{park2025anybcq} expands base quantization via per-precision scaling factors, and MatQuant \citep{nair2025matryoshka} derives low-bit models by truncating higher-bit representations. 
However, these approaches remain limited in practice: they either rely on hardware-unfriendly vector quantization \citep{park2024any}, introduce additional overhead from per-precision calibration parameters \citep{park2025anybcq}, or require offline repacking when increasing precision \citep{nair2025matryoshka}. Moreover, they lack efficiency for runtime adaptation, as switching between precisions requires loading extra scaling factors and launching distinct kernels (e.g., from 3-bit to 4-bit).
These limitations motivate our \texttt{MoBiQuant} framework, which mitigates precision-dependent outlier migration and enables efficient runtime precision switching without additional overhead.

\section{Motivation}
\label{sec:Motivation}
\textbf{Limitations of static PTQ.} 
Typical PTQ calibration process introduces a set of parameters $\mathbf{\Theta}_q$ to minimize the quantization error of the pretrained weights $\mathbf{W}$ at a target bit-width $b$. For example, OmniQuant~\cite{shao2023omniquant} learns to clip and rescale $\mathbf{W}$ and SpinQuant~\cite{liu2024spinquant} learns to rotate $\mathbf{W}$ to reduce outliers, respectively. 
Given a calibration set $\mathcal{X} = \{ \mathbf{X}_i \}_{i=1}^{N}$, where each sequence $\mathbf{X}_i \in \mathbb{R}^{T \times d}$ consists of $T$ tokens, the calibration process for static PTQ methods can be formulated as
\begin{equation}
\argmin_{\mathbf{\Theta}_q} \mathbb{E}_{\mathbf{X} \sim \mathcal{X}}  \mathcal{D} \big(\mathbf{X}, Q(\mathbf{W} | \mathbf{\Theta}_q, b)\big),
\label{PTQ}
\end{equation}
where $Q(\cdot)$ represents the weight quantizer and $\mathcal{D}(\cdot)$ denotes the quantization error function, respectively. 

\begin{figure}[ht]
\centering
\includegraphics[width=.95\linewidth]{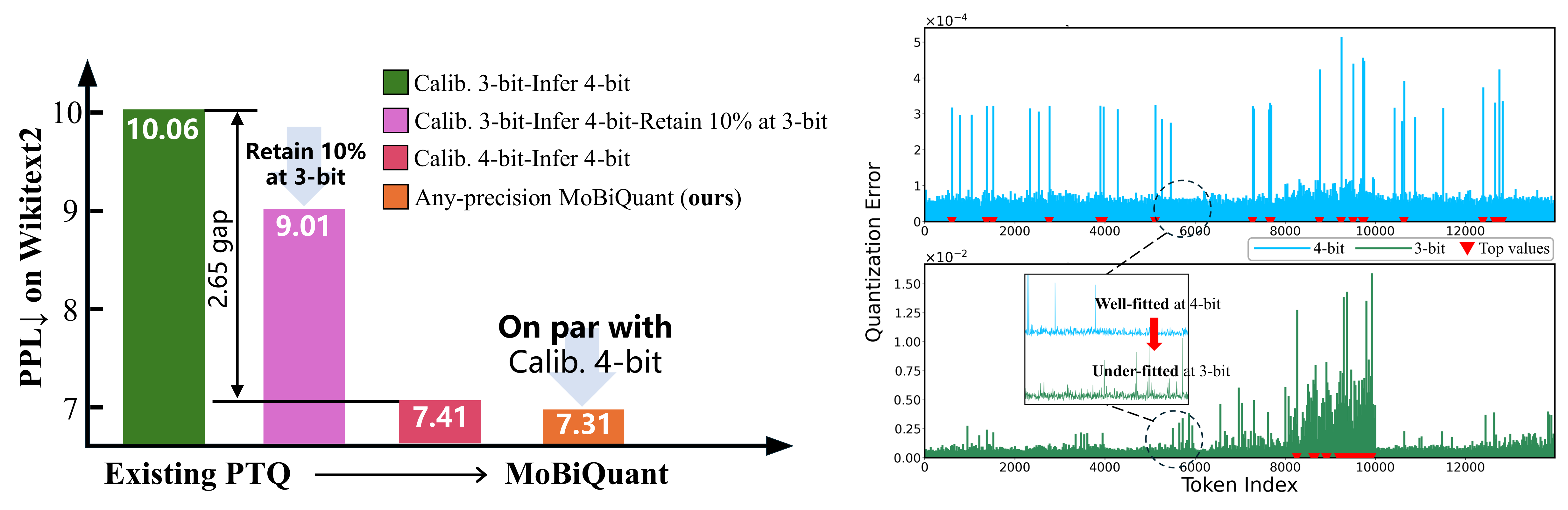}

\caption{(Left) Existing PTQ methods poorly generalize under calibration–inference precision mismatch, e.g., inference at 4-bit with 3-bit calibrated parameters (\textcolor{green!50!black}{green} vs. \textcolor{red!90!black}{red}). Incorporating token-aware bit adjustment partially mitigates this degradation (\textcolor[RGB]{216,110,204}{pink}). Our \texttt{MoBiQuant} improves generalization across precisions (\textcolor{red!50!yellow}{orange}). (Right) Per-token quantization error distribution at layer 5 in LLaMA3-8B: quantization outliers are highly non-uniform at each bit-width. Tokens well-fitted at 4-bit can become outliers at 3-bit, while 4-bit outliers may not be the primary error source at 3-bit.}
\label{motivation1}
\vspace{-5pt}
\end{figure}


The Eq.~(\ref{PTQ}) objective causes the parameters $\mathbf{\Theta}_q$ to overfit for the selected target bit-width $b$. We illustrate this limitation in Fig.~\ref{motivation1} (left): if we use the calibration parameters of 3-bit PTQ to quantize the LLaMA-3-8B model for 4-bit inference, the perplexity increases by $2.65$ points compared to a 4-bit calibrated model. This clearly highlights the lack of generalization across quantization precisions.




\textbf{Analysis of the poor generalization.}
Previous work \citep{chen2024prefixquant} has shown that the token sensitivity is highly non-uniform. Consequently, the calibration process optimizes $\mathbf{\Theta}_q$ to specifically fit the distribution of sensitive tokens using the overall objective in (\ref{PTQ}).  
We pinpoint the poor generalization of calibration parameters to the distribution shift of outlier tokens. We plot the per-token quantization error distributions when the target weight precisions are 3/4-bit for the same query in Fig.~\ref{motivation1} (right) and observe very different distributions.
Specifically, tokens that are accurately represented by 4-bit weights may emerge as dominant outliers for 3-bit weights, whereas outliers with 4-bit precision may no longer be the primary source of quantization error for 3-bit PTQ. We define this phenomenon as \textit{``outlier migration''} and it appears universally across different PTQ calibration methods either gradient-based OmniQuant~\cite{shao2023omniquant} or Hessian-based AWQ~\cite{lin2024awq} (see Appendices~\ref{app:AWQ} through \ref{app:quarot}).

To further motivate our approach, we revisit the 3-bit calibrated 4-bit inference experiment, but now retain the top 10\% of token outliers in the 3-bit calibration to be inferred with 3-bit weights. Counterintuitively, it shows that inferring tokens at a lower precision can yield higher overall performance (\textcolor[RGB]{216,110,204}{pink} bar). This observation suggests a promising direction: \textit{\textbf{token-adaptive precision adjustment} can improve the generalization of PTQ calibration parameters across bit-widths.}

\section{The MoBiQuant Framework}
\begin{figure*}[htb]
\centering
\includegraphics[width=.9\linewidth]{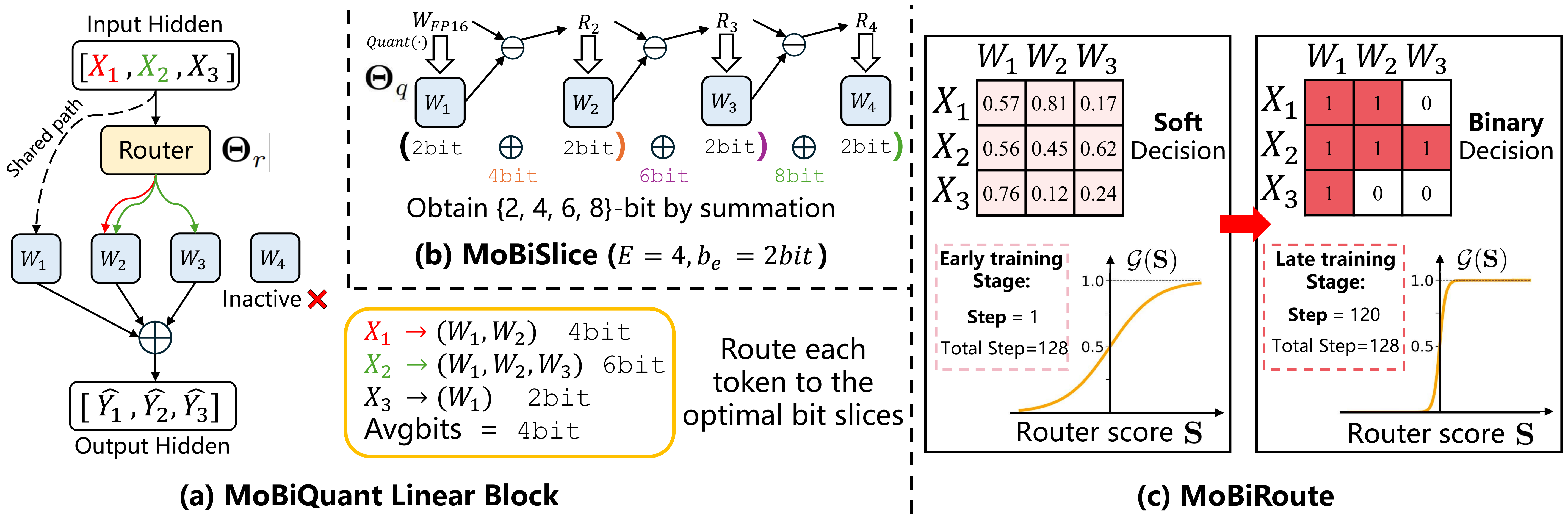}
\vspace{-7pt}
\caption{\texttt{MoBiQuant} enables token-adaptive any-precision inference for linear blocks in LLMs (a). It consists of two main components: \texttt{MoBiSlice} that supports multiple precisions by decomposing FP16 weights into bit slices via recursive quantization (b); and \texttt{MoBiRoute} with token-adaptive precision adjustment that is trained to route each token to the optimal bit slice under a budget (c).}
\label{fig:mobiquant}
\end{figure*}

Inspired by our previous observations, we present \texttt{MoBiQuant}, an any-precision quantization framework that mitigates \textit{outlier migration} through token-aware bit-width adaptation while enabling efficient runtime precision switching. An overview of the framework is shown in Fig.~\ref{fig:mobiquant}. 


\subsection{Bit Slicing via Recursive Quantization}\label{MoBiSliceSection}
Achieving token-level granularity requires a model to efficiently support multiple quantization levels (\eg 2--8-bit range). A na\"ive solution with separate models for each precision incurs high memory overhead. Existing any-precision approaches \citep{park2024any, wang2025d2moe, nair2025matryoshka} explicitly derive multiple precision levels from a single seed model by weight slicing or scaling. However, each precision remains a monolithic unit: switching a token to a different precision requires an offline reloading of the seed model and re-slicing, followed by redeployment. This is ill-suited for latency-sensitive settings where all adaptation is expected to occur at runtime.



To address this limitation, we propose a novel many-in-one quantization scheme \texttt{MoBiSlice}, where multiple precisions can be efficiently obtained at runtime. 
Our \texttt{MoBiSlice} decomposes the weight matrix $\mathbf{W}$ of each linear layer in the LLM into $E$ slices $\{\mathbf{W}_1, \mathbf{W}_2, \ldots, \mathbf{W}_E\}$, each containing a slice of $b_e$ bits from the quantized weight. This decomposition is implemented by recursively quantizing the residuals as
\begin{equation}
\mathbf{R}_1 = \mathbf{W},~
\mathbf{W}_e = Q(\mathbf{R}_e \mid \mathbf{\Theta}_q, b_e),~\textrm{and}~ 
\mathbf{R}_{e+1} = \mathbf{R}_e - \mathbf{W}_e,~\forall e \geq 1, 
\label{MoBiSliceeq}
\end{equation}
where the first shared most significant bit (MSB) slice $\mathbf{W}_1$ is obtained by quantizing the original $\mathbf{W}$ and the remaining slices $\mathbf{W}_{e>1}$ quantize the residuals $\mathbf{R}_e$ between the weight and existing slices in a recursive fashion. All slices are quantized using scalings derived from the shared $\mathbf{\Theta}_q$, where the scaling factor of the $(e+1)$-th slice is obtained by dividing that of the $e$-th slice by $2^{b_e - 1}$. This avoids the training overhead of multiple sets of scaling factors in existing methods \citep{park2024any,park2025anybcq}. Moreover, dequantization can be performed via efficient shift-and-add operations (Sec. \ref{sec:kernel}). Consequently, a weight $\mathbf{W}^{(b)}$ at a target precision $b$ can be constructed by summing a selection of $k$ bit slices as
\begin{equation}\label{eq:recon}
\mathbf{W}^{(b)} = \sum\nolimits_{e=1}^{k} \mathbf{W}_e,~\text{where}~b = \sum\nolimits_{e=1}^{k} b_e.
\end{equation}

In this paper, we adopt a default configuration with $E=4$ slices, each quantized to 2-bit, as shown in Fig.~\ref{fig:mobiquant} (b). In this way, weights with 4-bit and 6-bit precision can be constructed using 2 and 3 slices, respectively. This design enables any-precision inference using a fixed 2-bit kernel, thereby avoiding kernel relaunch overhead. In practice, \texttt{MoBiQuant} naturally learns to select the prefix $2k$-bit slices for inference without explicit constraints; further details are provided in Appendix~\ref{ap:mobislice}. \texttt{MoBiSlice} can be paired with existing PTQ pipelines, and we use OmniQuant~\citep{shao2023omniquant} as our PTQ backbone.  

\subsection{Runtime Precision Switching via Learned Token Routing}
\label{MoBiRouteSection}
Given bit slices provided by \texttt{MoBiSlice}, we propose a lightweight \texttt{MoBiRoute} router that dynamically assigns a different number of slices to each token with binary decisions. Specifically, given a sequence of $T$ tokens $\mathbf{X} \sim \mathcal{X}$, the routing score matrix $\mathbf{S} \in \mathbb{R}^{T \times E}$ is obtained as
\begin{equation}\label{eq:router}
\mathbf{S} = \mathcal{R}(\mathbf{X}, \mathbf{\Theta}_r), 
\end{equation}
where $\mathcal{R}(\cdot)$ is a learnable 2-layer MLP with the router parameters $\mathbf{\Theta}_r$, and each element of the score matrix $\mathbf{S}_{i,j}$ represents the confidence that the $i$-th token will require the $j$-th slice. 

Two key challenges exist when training the router in Eq.~(\ref{eq:router}):
\paragraph{Challenge 1: Learning binary routing decisions.}  
To determine whether a token $i$ should be routed to the $j$-th slice, one simple approach is to adopt a threshold-based routing strategy. However, fixed thresholds cannot adapt to a varying score distribution during the training stage. 
Therefore, our \texttt{MoBiRoute} learns a decision boundary implicitly using a differentiable gating function $\mathcal{G}(\cdot)$. When training is complete, $\mathcal{G}(\mathbf{S})$ acts as a learned binary mask applied to tokens. We formulate the proposed gating function as
\begin{equation} \mathcal{G}(\mathbf{S}) = \sigma \left(\tau(t) \cdot \mathbf{S}\right),~ \tau(t) = \ln(L) / \left(\ln(L) - \ln(t)\right), 
\label{eq:gating} 
\end{equation}
where $\sigma$ is the conventional sigmoid, and $\tau$ is the temperature defined by the current training step $t$ and the total number of steps $L$.

At the final step, the temperature $\tau(L)=\infty$, effectively converting the continuous gating function into a binary mask $\mathcal{G}(\mathbf{S}) = \mathbb{I}\left(\mathbf{S} > 0\right)$. The change in scores and the gating function during training are illustrated in Fig.~\ref{fig:mobiquant} (c).
Using the learned mask, the input token $\mathbf{X}_i$ is routed through the selected slices, and the corresponding outputs are aggregated to produce the output token $\hat{\mathbf{Y}}_i$ as
\begin{equation}
\hat{\mathbf{Y}}_i
= \sum\nolimits_{e=1}^{E}
\mathbf{W}_e^{\top}
\left(
\mathcal{G}(\mathbf{S})_{i,e} \odot \mathbf{X}_i
\right),
\label{eq:distribute}
\end{equation}
where input tokens are gated by the router's binary scores, and $\mathbf{W}_e$ is the bit slice with $b_e$ bits.

\paragraph{Challenge 2: Generalizing across precisions.} 
Any-precision inference requires the router to effectively accommodate a dynamic target precision budget (i.e., average bit-width). To achieve this, we apply a \textit{target precision scheduling} mechanism in the training process to encourage the router to explore different precision combinations. Specifically, we control router convergence through a budget-aware regularization objective defined by
\begin{equation}
\mathcal{L}_{\mathrm{reg}}(t) = \left( \operatorname{AvgBits} - b(t) \right) \cdot \left\lVert \mathcal{G}(\mathbf{S}) \right\rVert_1 ,~ 
b(t) = b_{\mathrm{init}} - \left( b_{\mathrm{init}} - b \right) \cdot \ln(t)/\ln(L),
\label{eq:reg}
\end{equation}
where the term $(\operatorname{AvgBits} - b(t))$ adaptively encourages bit slice pruning when the current average bit-width exceeds the target, and promotes slice activation otherwise.
The $\operatorname{AvgBits}$ in (\ref{eq:reg}) is estimated during training by counting activated bit slices per token. We define entries with scores exceeding $0.5$ in $\mathcal{G}(\mathbf{S})$ as active, which can be formulated as
\begin{equation}
\operatorname{AvgBits} = \frac{1}{T} \sum\nolimits_{i=1}^{T} \sum\nolimits_{j=1}^{E}
\mathbb{I}\!\left(\mathcal{G}(\mathbf{S})_{i,j} > 0.5\right) b_j,
\end{equation}
where $b_j$ denotes the bit slice size (e.g., 2 bits).
The scheduling term $b(t)$ starts with a higher precision $b_{\mathrm{init}}$ (\eg 8-bit) and gradually decays to the target $b$. This forces the router to explore a wide spectrum of token-slice assignments before converging to binary routing decisions. During training, the target $b$ is a hyperparameter and is set to 3-bit by default in our experiments. While \texttt{MoBiQuant} remains robust across different target precisions, varying the target bit-width may introduce minor performance variations, which we analyze in Appendix~\ref{app:target}.

\paragraph{Joint optimization.}
\label{alternatingraining}
Our framework consists of two sets of trainable parameters: $\mathbf{\Theta}_q$ and $\mathbf{\Theta}_r$ for \texttt{MoBiSlice} and \texttt{MoBiRoute}, respectively. They are jointly optimized during the PTQ process by
\begin{equation}
\label{equ:train_objective}
\argmin_{\mathbf{\Theta}_{q,r}} \mathbb{E}_{\mathbf{X} \sim \mathcal{X}} \left[ \mathcal{D} \left( \mathbf{W}^\top \mathbf{X}, \hat{\mathbf{Y}} | \mathbf{\Theta}_{q,r} \right) + \lambda \, \mathcal{L}_{\mathrm{reg}} \right],
\end{equation}
where $\mathcal{D}$ is the $L_2$ quantization error \ie~$\|\mathbf{Y} - \hat{\mathbf{Y}} \|_2$.
We freeze all weights from the pretrained LLM and calibrate only $\mathbf{\Theta}_q$ and $\mathbf{\Theta}_r$. We also treat $\mathbf{W}_1$ as a \textit{shared-expert slice} such that tokens always pass through for stable training.
We replace all linear layers in LLM transformer blocks with the proposed \texttt{MoBiQuant} block. Furthermore, we adopt a layer-wise calibration strategy from \citep{shao2023omniquant}. A detailed calibration strategy is presented in the Appendix~\ref{ap:alg}.
\paragraph{Efficient runtime precision switching.}
At the end of training, 
the score is away from zero if the router is confident about selection or exclusion of a bit slice, whereas the score remains at 0 with the lack of confident decision. 
Importantly, this separation allows us to achieve arbitrary precision adjustment at runtime by shifting the threshold of the score conversion process. During inference, we convert the router score into a binary selection mask using an adjustable threshold $\delta$ as
\begin{equation}
    \mathcal{G}_{\delta}(\mathbf{S}) = \mathbb{I}\left((\mathbf{S} - \delta) > 0\right),
\end{equation}
where $\delta$ can be globally adjusted for all layers at runtime.
Increasing $\delta$ reduces the number of activated slices per token, thereby lowering the effective precision, and vice versa. This design enables \texttt{MoBiQuant} to adapt precision at runtime without repeated weight repacking or dequantization.
\begin{figure}[htb]
\centering
\includegraphics[width=0.85\linewidth]{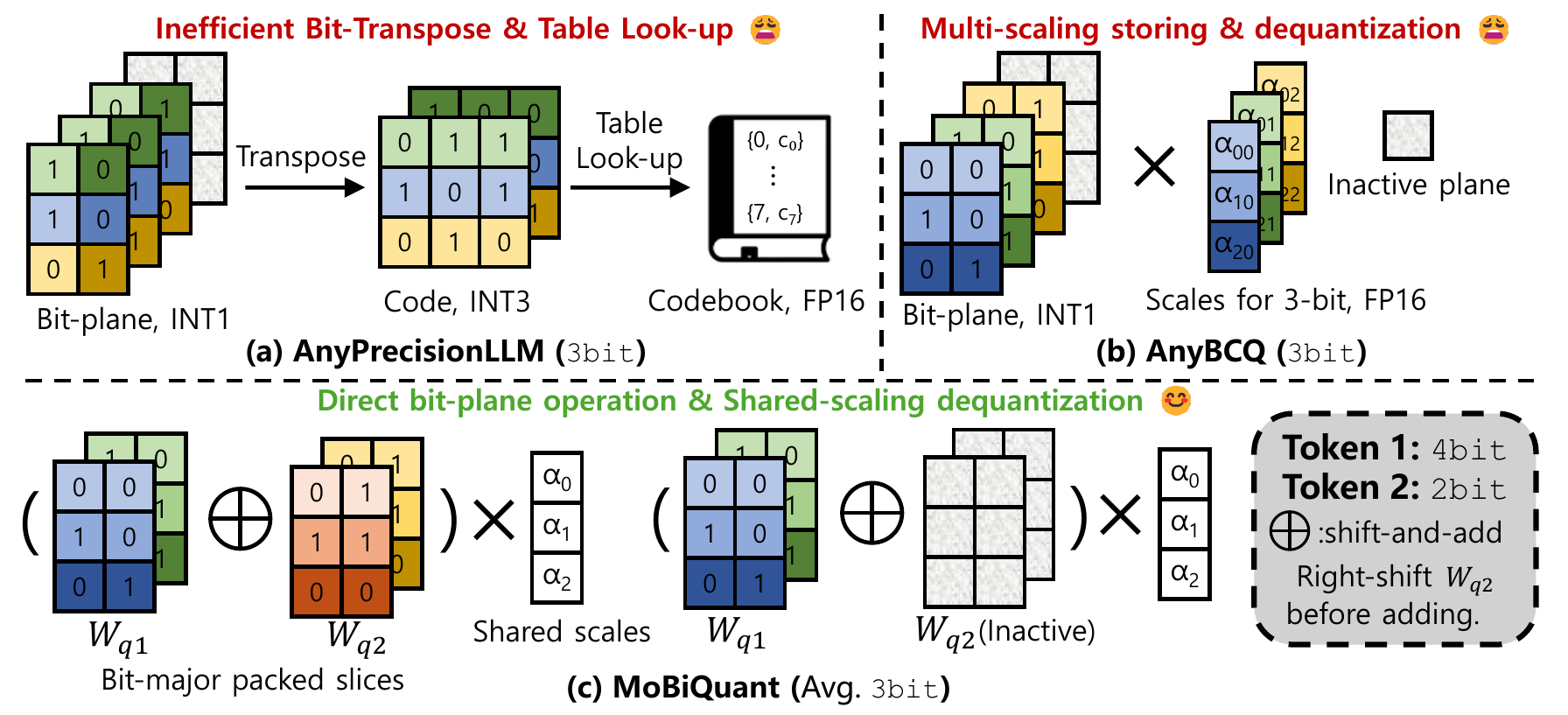}
\vspace{-7pt}
\caption{(a) AnyPrecisionLLM~\cite{park2024any} with the inefficient bit-plane transposes and table lookups, (b) AnyBCQ~\cite{park2025anybcq} with the multi-scale storage and dequantization overhead, and (c) our \texttt{MoBiQuant} that performs direct bit-plane operations with the shared scales for efficient any-precision dequantization.}
\vspace{-13pt}
\label{system}
\end{figure}
\subsection{MoBiQuant Kernel Design}
\label{sec:kernel}
The \texttt{MoBiQuant} kernel in Fig.~\ref{system} enables efficient inference by addressing the following challenges:

\textbf{\circnum{1} Redundant Memory Access.} 
Conventional static low-bit kernels typically load all weight slices regardless of the runtime precision, resulting in unnecessary memory traffic and limiting inference speedups. To enable on-demand fetching, we adopt the bit-major packing representation from~\citep{park2024any, park2025anybcq}, which decomposes quantized weights into independent bit-planes and stores each bit-plane in a separate contiguous memory region. The kernel can therefore independently fetch only the bit-planes required by the selected runtime precision. Within each active bit-plane, warp threads perform aligned and fully coalesced accesses, ensuring that skipping inactive bit-planes reduces the actual amount of data transferred from global memory. Tab.~\ref{tab:dram_traffic} reports the measured DRAM read traffic of a linear projection ($4096 \times 11008$) operation using the \texttt{MoBiQuant} kernel at different runtime precisions. The traffic decreases approximately proportionally as the bit-width is reduced, confirming that inactive bit-planes are not fetched from global memory.

\textbf{\circnum{2} Non-contiguous Memory Access and Load Imbalance.} 
During prefilling and batched decoding, tokens are dynamically assigned to different bit slices, requiring gathers from non-contiguous memory locations and reducing memory bandwidth utilization. To address this, we permute tokens after routing so that tokens assigned to the same bit slice are stored contiguously. Tab.~\ref{tab:permutation} quantifies the permutation overhead and the resulting prefilling latency reduction for a 4-bit MLP layer in LLaMA-2-7B, showing that improved memory locality and load balancing outweigh the permutation overhead and yield a net speedup.



\begin{table}[htbp]
\centering
\vspace{-7pt}
\begin{minipage}[t]{0.47\textwidth}
    \centering
    \small
    \caption{DRAM read traffic across bit-wdiths.}
    \label{tab:dram_traffic}
    
    \begin{tabular}{c c c c}
        \toprule
        \textbf{Bit} & \textbf{Planes} &
        \textbf{DRAM Read} & \textbf{$\Delta$/Plane} \\
        \midrule
        2 & 1 & 12.73 MB & -- \\
        4 & 2 & 24.05 MB & +11.32 MB \\
        6 & 3 & 35.40 MB & +11.35 MB \\
        8 & 4 & 46.68 MB & +11.28 MB \\
        \bottomrule
    \end{tabular}
\end{minipage}
\hfill
\begin{minipage}[t]{0.5\textwidth}
    \centering
    \small
    \caption{Contribution of token permutation. }
    \label{tab:permutation}
    
    \begin{tabular}{c c c c}
        \toprule
        \textbf{Length} & \textbf{Method} &
        \textbf{Perm. (ms)} & \textbf{Total (ms)} \\
        \midrule
        2048 & w/o perm. & -- & 5.857 \\
             & w/ perm.  & 0.134 (2.9\%) & \textbf{4.609} \\
        \midrule
        4096 & w/o perm. & -- & 9.916 \\
             & w/ perm.  & 0.179 (2.3\%) & \textbf{7.954} \\
        \bottomrule
    \end{tabular}
\end{minipage}
\vspace{-7pt}
\end{table}

\textbf{\circnum{3} Routing Overhead.}
\texttt{MoBiQuant} improves kernel efficiency in two aspects: First, unlike AnyPrecisionLLM~\citep{park2024any}, \texttt{MoBiQuant} avoids bit transposition and centroid lookup by performing Binary Matrix Multiplication (BMMA) directly on packed bit-planes. Second, compared with AnyBCQ~\citep{park2025anybcq}, \texttt{MoBiQuant} shares scaling factors across slices and combines slices at the bit-plane level before scaling, reducing dequantization overhead (Fig.~\ref{system}).
However, \texttt{MoBiQuant} introduces token routing. To mitigate the additional routing overhead, we fuse all router projections within each layer into a single GPU kernel with shared-memory input reuse. Fig.~\ref{kernelmeasurements} reports the prefill latency breakdown and end-to-end batched-decoding performance of the three methods.

\section{Experiments}
\label{Section5}
\subsection{Experimental Setup}
\paragraph{Models and benchmarks.} To cover a broad range of model scales, we conduct experiments on LLaMA2-7B/13B~\citep{touvron2023llama}, LLaMA3-8B, LLaMA3.2-1B/3B, Gemma2-9B~\citep{team2024gemma}, and Qwen3-32B~\citep{yang2025qwen3}. We evaluate perplexity on WikiText2~\citep{merity2016pointer} and accuracy on MMLU~\citep{hendrycks2020measuring}. We further report zero-shot performance on BoolQ~\citep{clark2019boolq}, PIQA~\citep{Bisk2020}, HellaSwag~\citep{zellers2019hellaswag}, WinoGrande~\citep{sakaguchi2021winogrande}, ARC-Easy, and ARC-Challenge~\citep{clark2018think}, as well as reasoning performance on GSM8K~\citep{cobbe2021training}.

\textbf{Baselines and implementation.}
We primarily compare \texttt{MoBiQuant} with the any-precision baselines AnyPrecisionLLM~\citep{park2024any} and AnyBCQ~\citep{park2025anybcq}. We further compare against single-precision PTQ methods, including GPTQ~\citep{frantar2022gptq}, OmniQuant~\citep{shao2023omniquant}, AWQ~\citep{lin2024awq}, SmoothQuant~\citep{xiao2023smoothquant}, SpinQuant~\citep{liu2024spinquant}, and QuaRot~\citep{ashkboos2024quarot}, as well as the VQ-based methods QUIP\#~\citep{tseng2024quip} and QTIP~\citep{tseng2024qtip}. We use 128 sequences from WikiText2 for calibration, and measure all kernel results on NVIDIA A100 GPUs with CUDA 12.9. Additional experimental details are provided in Appendix~\ref{app:exp_detail}.
\begin{figure*}[htb]
\centering
\includegraphics[width=.95\linewidth]{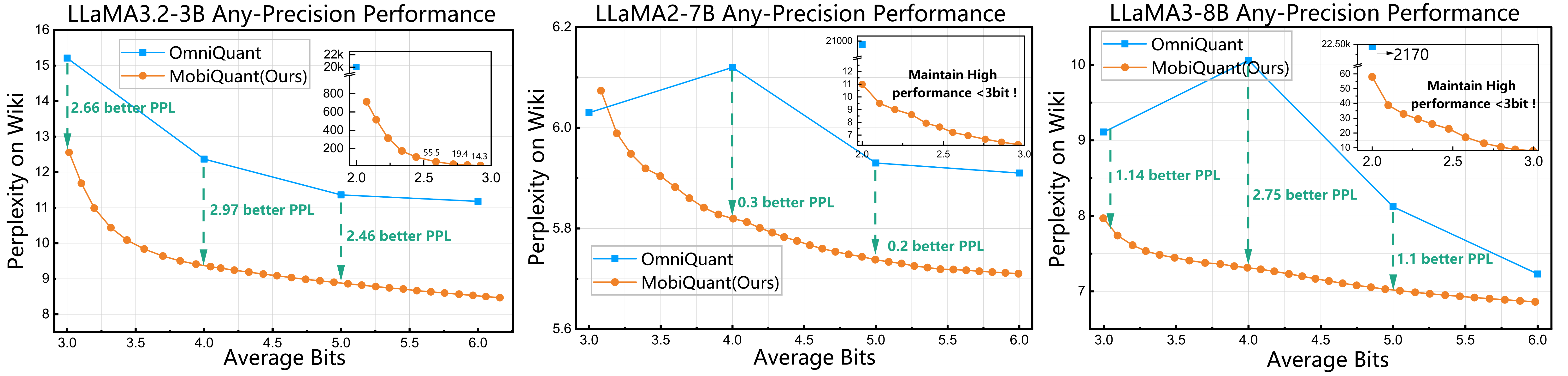}
\vspace{-7pt}
\caption{\textbf{Any-precision inference performance comparison.} \texttt{MoBiQuant} outperforms the PTQ baseline across multiple unseen target bit-widths on various LLaMA models, maintaining smooth and fine-grained precision scaling even at extremely low 2--3-bit precision range.}
\vspace{-14pt}
\label{elasticperformance}
\end{figure*}
\subsection{Main Results} 
\textbf{Improved Cross-Bit Generalization.} We first investigate whether \texttt{MoBiQuant} improves the cross-bit generalization of our integrated PTQ baseline, OmniQuant~\cite{shao2023omniquant}.
 Specifically, we set the calibration bit-width for both OmniQuant and \texttt{MoBiQuant} to 3-bit and evaluate their performance across unseen higher and lower bit-widths. 
As shown in Fig.~\ref{elasticperformance},
\texttt{MoBiQuant} consistently outperforms OmniQuant at all unseen bit-widths across various models. Importantly, even at extremely low 2--3-bit range, \texttt{MoBiQuant} maintains smooth and stable precision scaling, whereas OmniQuant exhibits significant performance degradation. We show that \texttt{MoBiQuant} can also improve rotation-based PTQ methods (Appendix~\ref{app:quarot}). Similar trend is also observed in activation quantization (Appendix~\ref{app:activation}).

\textbf{Comparison with any-precision baselines.}
We compare \texttt{MoBiQuant} with AnyPrecisionLLM~\citep{park2024any} and AnyBCQ~\citep{park2025anybcq} in Tab.~\ref{tab:anyprecision} in terms of perplexity, accuracy, and decoding throughput. On LLaMA2-7B at 2-bit, \texttt{MoBiQuant} reduces WikiText2 perplexity from $2\mathrm{e}3$ and $15.38$ to $10.91$, while improving MMLU accuracy from $25.11\%$ and $29.87\%$ to $33.20\%$. The improvements generalize to larger models Gemma2-9B and Qwen3-32B. Similar advantages are observed at 3- and 4-bit, where \texttt{MoBiQuant} generally provides a favorable accuracy--precision trade-off. Moreover, by avoiding centroid lookup and multi-scalar dequantization, the \texttt{MoBiQuant} kernel achieves an average decoding-throughput improvement of $33.8\%$ over AnyPrecisionLLM and $22.8\%$ over AnyBCQ on the LLaMA models. We also compare the calibration cost in Appendix~\ref{app:calibration_cost}.

\begin{table}[htb]
\vspace{-5pt}
\centering
\scriptsize
\caption{End-to-end comparison with the any-precision baselines AnyPrecisionLLM (AP) and AnyBCQ (ABC). \texttt{MoBiQuant} achieves higher decoding throughput while maintaining competitive perplexity and accuracy.}
\label{tab:anyprecision}
\begin{tabular}{cc|cccc|cccc|ccc}
\specialrule{0.8pt}{0pt}{0pt}
\multirow{2}{*}{\textbf{Model}} &
\multirow{2}{*}{\textbf{Bits}} &
\multicolumn{4}{c|}{\textbf{WikiText2, PPL $\downarrow$}} &
\multicolumn{4}{c|}{\textbf{MMLU, Acc. $\uparrow$}} &
\multicolumn{3}{c}{\textbf{Throughput, tokens/sec. $\uparrow$}} \\
\cline{3-13}
& & FP16 & AP & ABC & MoBiQ
  & FP16 & AP & ABC & MoBiQ
  & AP & ABC & MoBiQ \\
\hline
\multirow{3}{*}{LLaMA2-7B}
& 2 & 5.47 & 2e3 & 15.38 & \textbf{10.91}
    & 45.30 & 25.11 & 29.87 & \textbf{33.20}
    & 279 & 312 & \textbf{404} \\
& 3 & 5.47 & 6.13 & 6.25 & \textbf{6.07}
    & 45.30 & 41.85 & 41.56 & \textbf{43.35}
    & 241 & 268 & \textbf{311} \\
& 4 & 5.47 & 6.05 & 5.91 & \textbf{5.82}
    & 45.30 & 41.74 & \textbf{43.84} & 43.28
    & 211 & 225 & \textbf{256} \\
\hline
\multirow{3}{*}{LLaMA3-8B}
& 2 & 6.12 & 1e3 & \textbf{19.01} & 58.12
    & 65.02 & 24.66 & 35.32 & \textbf{40.13}
    & 228 & 245 & \textbf{352} \\
& 3 & 6.12 & 8.60 & 8.08 & \textbf{7.97}
    & 65.02 & 55.53 & 58.28 & \textbf{59.71}
    & 194 & 212 & \textbf{257} \\
& 4 & 6.12 & \textbf{6.70} & 6.84 & 7.31
    & 65.02 & \textbf{64.04} & 63.15 & 63.29
    & 171 & 186 & \textbf{209} \\
\hline
\multirow{3}{*}{Gemma2-9B}
& 2 & 6.84 & 19.62 & 12.72 & \textbf{12.21}
    & 70.74 & 33.09 & 43.36 & \textbf{44.28}
    & 163 & 185 & \textbf{266} \\
& 3 & 6.84 & 7.94 & 7.95 & \textbf{7.73}
    & 70.74 & 65.30 & 65.38 & \textbf{66.41}
    & 144 & 164 & \textbf{199} \\
& 4 & 6.84 & \textbf{7.06} & 7.23 & 7.07
    & 70.74 & 69.23 & 68.81 & \textbf{69.08}
    & 125 & 146 & \textbf{164} \\
\hline
\multirow{3}{*}{Qwen3-32B}
& 2 & 7.61 & 12.58 & 12.07 & \textbf{11.50}
    & 81.85 & 71.23 & 70.70 & \textbf{72.68}
    & 43 & 52 & \textbf{75} \\
& 3 & 7.61 & 8.93 & 8.42 & \textbf{8.14}
    & 81.85 & 78.95 & 79.72 & \textbf{80.43}
    & 40 & 46 & \textbf{56} \\
& 4 & 7.61 & 7.87 & 7.81 & \textbf{7.75}
    & 81.85 & 80.82 & 80.70 & \textbf{80.97}
    & 35 & 42 & \textbf{45} \\
\specialrule{0.8pt}{0pt}{0pt}
\end{tabular}
\vspace{-5pt}
\end{table}

\textbf{Comparison with single-precision PTQ baselines.}
We evaluate prior static scalar PTQ baselines that have been separately trained for fixed 3- and 4-bit precisions. For fair comparison, we keep our \texttt{MoBiQuant} elastic during inference but restrict its average precision to be either 3- or 4-bit.
As shown in Tab.~\ref{tab:staticcomparison}, \texttt{MoBiQuant} consistently matches the performance of state-of-the-art PTQ methods on LLaMA2 models and even surpasses them on LLaMA-3 models. 
Quantitative results demonstrate that \texttt{MoBiQuant} attains the performance of static quantization methods while preserving flexible precision adjustment. Additional zero-shot and GSM8K results, along with comparisons against VQ-based methods, are provided in Appendix~\ref{app:staticptq}.

\newcommand{\NR}{\textcolor{gray}{\scriptsize\textit{NR}}}
\newcommand{\NRone}{\multicolumn{1}{c}{\NR}}
\newcommand{\NRtwo}{\multicolumn{2}{c|}{\NR}}
\newcommand{\NRtwoNoBar}{\multicolumn{2}{c}{\NR}}

\newcommand{\NA}{\textcolor{gray}{\scriptsize\textemdash}}
\newcommand{\NAone}{\multicolumn{1}{c}{\NA}}
\newcommand{\NAtwo}{\multicolumn{2}{c|}{\NA}}
\newcommand{\NAtwoNoBar}{\multicolumn{2}{c}{\NA}}

\newcommand{\MISS}{\NR} 

\newcommand{\MISSone}{\multicolumn{1}{c}{\MISS}}
\newcommand{\MISStwo}{\multicolumn{2}{c|}{\MISS}}
\newcommand{\MISStwoNoBar}{\multicolumn{2}{c}{\MISS}}

\begin{table*}[htbp]
\captionsetup{justification=justified}
\caption{The proposed any-precision \texttt{MoBiQuant} \textbf{matches} or \textbf{surpasses} the static scalar quantization (SQ) PTQ baselines with the same number of average bits on WikiText2, PPL.
}
\vspace{-5pt}
\label{tab:staticcomparison}
\centering
\footnotesize
\begin{tabular}{c|c|cc|cc|cc|cc|cc}
\toprule
\multirow{2}{*}{\textbf{Method}} &
\multirow{2}{*}{\textbf{Type}} &
\multicolumn{2}{c|}{LLaMA2-7B} &
\multicolumn{2}{c|}{LLaMA2-13B} &
\multicolumn{2}{c|}{LLaMA3.2-1B} &
\multicolumn{2}{c|}{LLaMA3.2-3B} &
\multicolumn{2}{c}{LLaMA3-8B} \\
& & 3-bit & 4-bit & 3-bit & 4-bit & 3-bit & 4-bit & 3-bit & 4-bit & 3-bit & 4-bit \\
\midrule
FP16 & \NA &
\multicolumn{2}{c|}{5.5} &
\multicolumn{2}{c|}{5.0} &
\multicolumn{2}{c|}{13.4} &
\multicolumn{2}{c|}{9.2} &
\multicolumn{2}{c}{6.1} \\
\midrule
SmoothQuant & \multirow{6}{*}{\shortstack{Static\\SQ}} &
8.62 & 7.50 &
5.43 & 6.10 &
2e2 &2e2 &
3e2 &1e2 &
12.50 & 10.70 \\
AWQ & &
6.24 & 5.83 &
5.32 & 5.07 &
20.69 &15.02 &
15.51 &12.76 &
8.22 & 7.36 \\
GPTQ & &
6.29 & 5.90 &
5.42 & 5.60 &20.81
&15.19  &
15.77& 13.14 &
8.28 & 7.43 \\
SpinQuant & &
6.13 & \textbf{5.60} &
5.24 & \textbf{5.00} &
19.62 &14.45 &
\textbf{14.73} &11.17 &
8.14 & \textbf{6.40} \\
QuaRot  & &
6.13 &5.61  &
\textbf{5.22} &5.03  &
20.15 &14.77 &
14.82 & 11.55 &
9.01 & 7.20 \\
OmniQuant & &
\textbf{6.03} & 5.74 &
5.28 & 5.02 &
23.92 & 16.79 &
15.21 & 12.71 &
9.11 & 7.41 \\
\midrule
MoBiQuant &
Any-prec. &
6.07 & 5.82 &
5.31 & 5.08 &
\textbf{19.40} & \textbf{14.02} &
14.76 & \textbf{9.41} &
\textbf{7.97} & 7.31 \\
\bottomrule

\end{tabular}
\vspace{-7pt}
\end{table*}

\subsection{Analytical Results}

\begin{figure}[h]
\centering
\includegraphics[width=.9\linewidth]{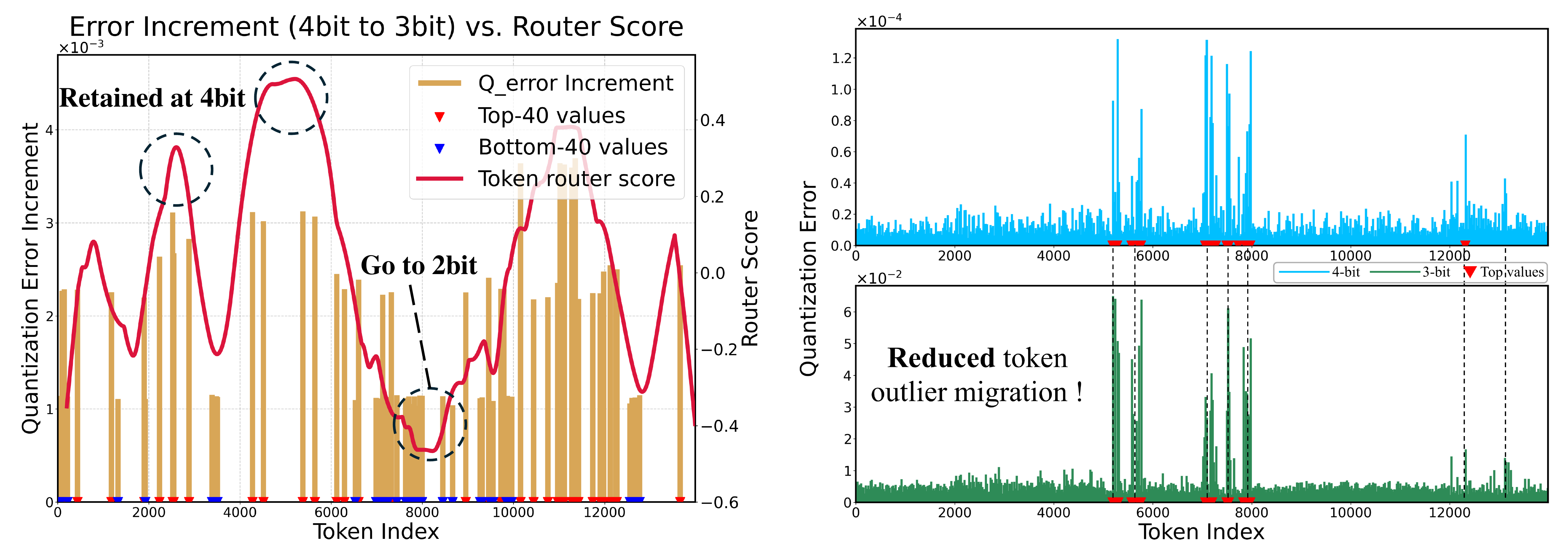}
\vspace{-7pt}
\caption{\textbf{Visualization of outlier migration mitigation.}  \texttt{MoBiQuant} router scores correlate with the actual token-wise error increments under precision switching (left), and it produces more consistent token-wise error distributions across bit-widths with reduced outlier migration (right).}
\vspace{-13pt}
\label{motivationverification}
\end{figure}

\textbf{Causal link between Token Routing and Outlier Migration.} 
As discussed in Section~\ref{sec:Motivation}, \textit{outlier migration} exposes a calibration--inference mismatch: tokens well preserved at one precision can become disproportionately sensitive at another.
To examine whether \texttt{MoBiQuant} mitigates this mismatch, we compare token-wise error increments when switching OmniQuant from 4-bit calibration to 3-bit inference with \texttt{MoBiRoute}'s router scores.
Fig.~\ref{motivationverification} (left) shows that tokens with larger error increments tend to receive higher router scores and higher precision, while less sensitive tokens are routed to 2-bit computation, indicating that the router captures sensitivity to precision reduction.
Fig.~\ref{motivationverification} (right) shows more consistent token-wise error distributions across bit-widths than static PTQ in Fig.~\ref{motivation1}, indicating reduced \textit{outlier migration}.
These observations support the proposed token-adaptive routing allocates higher precision to vulnerable tokens and mitigates the calibration--inference mismatch.

\begin{figure}[htb]
\centering
\vspace{-5PT}
\includegraphics[width=.9\linewidth]{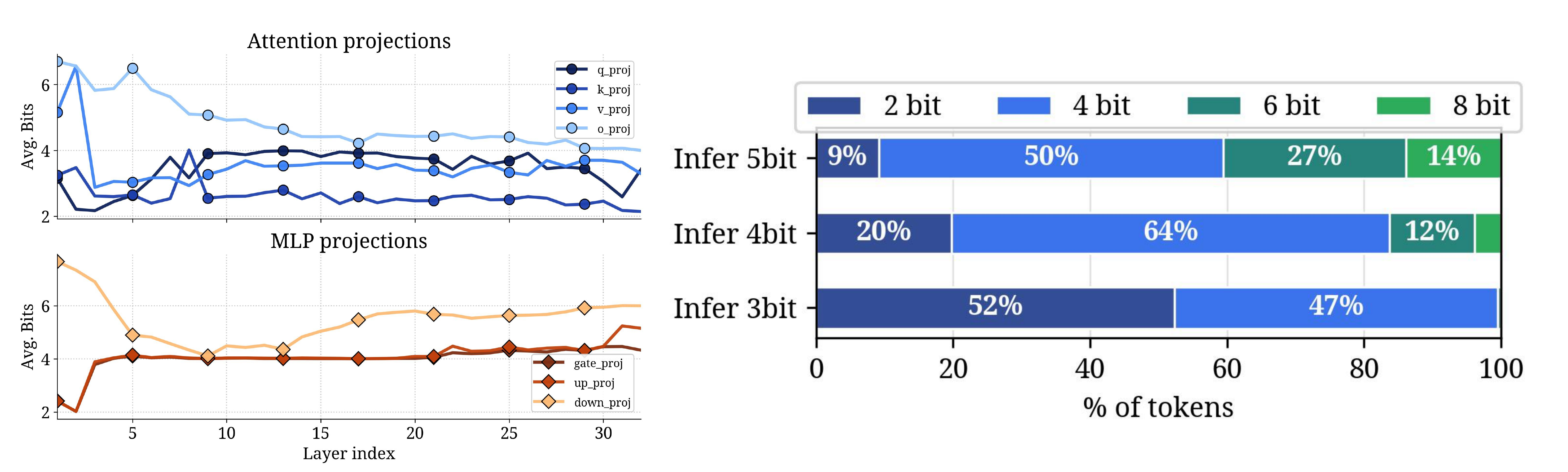}
\vspace{-7pt}
\caption{(Left) Average precision assignments of different linear blocks by \texttt{MoBiQuant}, showing block-level variation and higher deviations in initial layers. (Right) Precision distributions for any-precision inference, showing heterogeneous token assignments across target bit-widths.}
\label{TokenPrecision}
\vspace{-16pt}
\end{figure}

\textbf{Token- and block-wise assignments.}
We collect statistics on token bit-width assignments for any-precision inference. As shown in Fig.~\ref{TokenPrecision} (right), tokens have assignment distributions that shift with target precision \eg most tokens are assigned to 2- or 4-bit precisions under 3-bit inference, whereas the majority of tokens shift toward the 4--6-bit range for the 5-bit inference. Overall, the average  precision remains within the prescribed target  token-aware bit adjustment.
We also observe that the average precision varies across different linear blocks, even though each layer is trained for the same target bit-width. As shown in Fig.~\ref{TokenPrecision} (left), 
MLP blocks exhibit divergent assignments across layers, and the first layers tend to have deviations in bit-width assignments. This pattern is consistent with the \cite{hooper2025fgmp} study that allocates different precisions to blocks based on their sensitivity, indicating that \texttt{MoBiQuant} also benefits from block-level precision adjustments.

\subsection{Kernel Evaluation}

\label{kernel evaluation}
\begin{figure}[htb]
\centering
\vspace{-10pt}
\includegraphics[width=.9\linewidth]{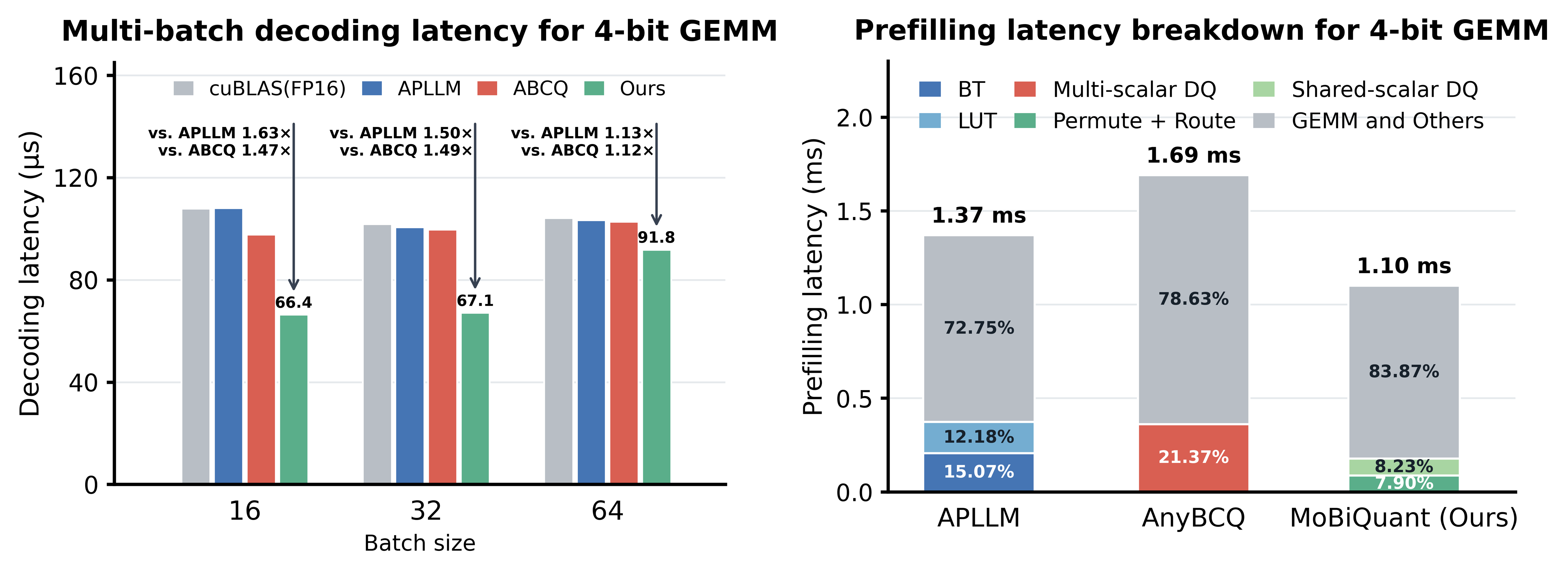}
\vspace{-12pt}
\caption{\textbf{Kernel evaluation of \texttt{MoBiQuant}.} (Left) Multi-batch decoding latency across different batch sizes. (Right) Prefill latency breakdown of any-precision methods.}
\label{kernelmeasurements}
\vspace{-17pt}
\end{figure}

We evaluate the \texttt{MoBiQuant} kernel under multi-batch decoding and prefilling workloads using a GEMM shape of $(14336 \times 4096)$. Figure~\ref{kernelmeasurements} (left) compares the latency of the 4-bit GEMM against cuBLAS FP16, AnyPrecisionLLM (APLLM)~\citep{park2024any}, and AnyBCQ (ABCQ)~\citep{park2025anybcq} across different decoding batch sizes. \texttt{MoBiQuant} consistently achieves the lowest latency, providing $1.13$--$1.63\times$ speedup over APLLM and $1.12$--$1.49\times$ over ABCQ for batch sizes ranging from 16 to 64. Figure~\ref{kernelmeasurements} (right) further reports the prefill latency breakdown with a context length of 2048. Despite introducing token routing and permutation, \texttt{MoBiQuant} limits their overhead to only $7.90\%$ of the total latency. Its shared-scalar dequantization also incurs less overhead than the bit-transposition and lookup-table operations in APLLM and the multi-scalar dequantization in ABCQ. Consequently, \texttt{MoBiQuant} achieves the lowest overall prefill latency. Additional results on memory savings are provided in Appendices~\ref{app:MemSavings}.
\subsection{Ablation Study}
We evaluate the sensitivity to the calibration data by replacing WikiText2 with two alternative datasets: C4~\citep{raffel2020exploring} and PTB~\citep{marcus1993building}. \texttt{MoBiQuant} consistently achieves better results in most calibration–evaluation pairs, demonstrating robustness to calibration set selection. The full results are shown in the Appendix~\ref{app:downstream}.
We also report ablation studies on router regularization scheduling in Appendix~\ref{app:schedule}, and examine router generalization for different target precision training in Appendix~\ref{app:target}. 


\section{Conclusion}\label{sec:conc}
In this paper, we identify the ``\textit{outlier migration}'' in conventional PTQ methods, which limits generalization of calibration across precisions. To address it, we proposed \texttt{MoBiQuant}, a quantization framework that integrates \texttt{MoBiSlice} for fine-grained precision selection and \texttt{MoBiRoute} for token-adaptive routing. Extensive experiments demonstrate its effectiveness and practicality.

\begin{ack}



This work was supported in part by NSF (No. 2625580), ARO (No. W911NF-23-2-0224), and DOE (No. DE-SC0026696).
Additional support was provided by the National Research Foundation of Korea (NRF) (No. RS-2025-02216217); the Institute of Information \& Communications Technology Planning \& Evaluation (IITP), funded by the Korean government (MSIT) (No. RS-2019-II190421, AI Graduate School Support Program (Sungkyunkwan University)); and the Korea Institute of Police Technology (KIPoT), funded by the Korean National Police Agency and Korea Customs Service (No. RS-2026-25536784).
This work was also supported in part by the National Research Foundation of Korea (NRF), funded by the Korean government (MSIT) (No. RS-2026-25499095).


\end{ack}

\bibliography{nips26}
\bibliographystyle{unsrt}  

\newpage
\appendix
\section*{Appendix}

\newcommand{\TOCdots}{\leaders\hbox{.\kern0.6pt}\hfill}

\newcommand{\TOCentry}[3][0em]{%
  \noindent\hspace*{#1}%
  \hyperref[#2]{#3}%
  \TOCdots
  \hyperref[#2]{\pageref{#2}}%
  \par
}

\newcommand{\TOClineAuto}[2]{\TOCentry[0em]{#2}{#1}}
\newcommand{\TOCsublineAuto}[2]{\TOCentry[1.5em]{#2}{#1}}

\section*{Contents}

\TOClineAuto{A. Overall Algorithm of MoBiQuant}{ap:alg}

\TOClineAuto{B. MoBiSlice Construction: Formulation and Further Analysis}{ap:mobislice}

\TOClineAuto{C. Experiment Details}{app:exp_detail}
\TOCsublineAuto{C.1. MoBiQuant Training Setup}{app:setup}
\TOCsublineAuto{C.2. Flexible Inference and Effective Bit Control}{app:flexible}
\TOCsublineAuto{C.3. Baseline Results}{app:baseline_results}
\TOClineAuto{D. Ablation Studies}{app:ablation}
\TOCsublineAuto{D.1. Ablation of Calibration Dataset}{app:downstream}
\TOCsublineAuto{D.2. Ablation of Scheduling Methods for Router Regularization}{app:schedule}
\TOCsublineAuto{D.3. Ablation of Different Training Target Bit}{app:target}

\TOClineAuto{E. Additional Experiments}{app:exp}
\TOCsublineAuto{E.1. Outlier migration in Non-Gradient-Based PTQ}{app:AWQ}
\TOCsublineAuto{E.2. Outlier migration on Mistral-7B}{app:mistral}
\TOCsublineAuto{E.3. Compatibility with rotation-based methods}{app:quarot}
\TOCsublineAuto{E.4. Any-precision performance with Activation Quantization}{app:activation}
\TOCsublineAuto{E.5. Additional Comparison with Static PTQ Methods}{app:staticptq}
\TOCsublineAuto{E.6. Memory Savings under Multi-Precision Deployment}{app:MemSavings}
\TOClineAuto{F. Limitations}{app:limitations}


\section{Overall Algorithm of MoBiQuant}
\label{ap:alg}
As shown in Alg.~\ref{alg:mobiquant_overall}, \texttt{MoBiQuant} performs layer-wise quantization by matching each quantized layer to a full precision reference output using a two-stage iterative process. For each layer, it first runs the original full-precision layer to obtain full precision outputs, then initializes the shared quantization parameters $\mathbf{\Theta_q}$ and the router $\mathbf{\Theta_r}$. In \textit{Stage 1}, it stabilizes the first slice path by optimizing the quantization parameters to minimize the reconstruction error to the full-precision target. In \textit{Stage 2}, it initializes residual slices' quantization parameters from \textit{Stage 1}, computes the router score for each token, and produces the overall quantized outputs. This stage jointly optimizes the objective that combines reconstruction loss with a router score regularization loss. After the optimization converges for the layer, \texttt{MoBiQuant} commits the learned quantization parameters, propagates both full-precision and quantized outputs to the next layer, and repeats the same procedure until all layers are quantized.

\section{MoBiSlice Construction: Formulation and Further Analysis}
\label{ap:mobislice}

In this section, we formalize the \texttt{MoBiSlice} quantization scheme and describe how we decompose a full-precision weight into a set of composable low-bit slices, enabling a many-in-one representation for multiple precisions.

\paragraph{Quantizer design for bit slice accumulation.}
\texttt{MoBiSlice} requires that precision be adjustable by either accumulating residual bits or truncating them. To satisfy this property, we adopt a floor-aligned mapping following the truncation-ready quantization principle \cite{kim2025truncquant}, where a lower precision code is obtained by simply dropping least significant bits (LSB) rather than re-rounding. Concretely, for bit-width $b$, scale $s$, and continuous zero point $z$, we quantize as
\begin{align}
& x_{\mathrm{int}}=\mathrm{clamp}\!\left(\left\lfloor \frac{x}{s}+z \right\rfloor,\,0,\,2^{b}-1\right), \\
& x_{\mathrm{deq}} = s\left(x_{\mathrm{int}}-z+0.5\right).
\end{align}
The floor operator enforces hierarchical nesting of integer codes, so switching bit width corresponds to adding or dropping bit slices without changing previously formed higher order bits. The $+0.5$ dequantization shift centers each bin, which reduces bias when residual slices are accumulated.

\definecolor{OursText}{RGB}{54,125,189}

\begin{algorithm}[H]
\caption{\textsc{MoBiQuant Calibration}}
\label{alg:mobiquant_overall}
\begin{algorithmic}[1]
\setlength{\itemsep}{0.2em}       

\Statex \textbf{Input:} Input tokens $\mathbf{X}$, model $\mathcal{M}$ with $L$ layers $\{f^{(\ell)}_{\mathrm{fp}}\}_{\ell=1}^{L}$,
        epochs \texttt{ep}, number of bit slices $E$, regularization weight $\lambda$
\Statex \textbf{Output:} Quantized model $\widehat{\mathcal{M}}$

\Statex \textbf{Notation:} We omit the layer superscript $(\ell)$ for readability.
\Statex \quad $\mathbf{H}_{\mathrm{fp}}, \mathbf{H}_{\mathrm{q}}$: Full precision and quantized input for layer $\ell$
\Statex \quad $\mathbf{Y}_{\mathrm{fp}}, \mathbf{Y}_{\mathrm{q}}$: Full precision and quantized output for layer $\ell$
\Statex \quad $\mathbf{G}$: Router scores, $\odot$: element wise product

\vspace{0.35em}
\State Initialize activations $\mathbf{H}_{\mathrm{fp}}, \mathbf{H}_{\mathrm{q}} \gets \mathbf{X}$

\For{$\ell = 1$ \textbf{to} $L$}

    \State FP output $\mathbf{Y}_{\mathrm{fp}} \gets f_{\mathrm{fp}}(\mathbf{H}_{\mathrm{fp}};\mathbf{W})$

    \State Initialize quantization parameters $\mathbf{\Theta_{q}}$ and router parameters $\mathbf{\Theta_{r}}$
    
    \For{$t = 1$ \textbf{to} \texttt{ep}}
        \Statex \color{OursText}{\textbf{\textit{\quad \quad \quad Stage 1: First slice (FS) stabilization}}}
        
        \State $\mathbf{Y}_{\mathrm{FS}} \gets f_{\mathrm{FS}}(\mathbf{H}_{\mathrm{q}};\mathbf{\Theta_{q}})$
        \hfill $\triangleright$ First slice-only forward pass
        
        \State $\mathcal{L}_{\mathrm{FS}} \gets \|\mathbf{Y}_{\mathrm{FS}} - \mathbf{Y}_{\mathrm{fp}}\|_2^2$
        \hfill $\triangleright$ match FP reference output
        
        \State Update $\mathbf{\Theta_{q}}$
        
        \Statex \color{OursText}{\textbf{\textit{\quad \quad \quad Stage 2: Joint training}}}
        
        \State Slices $\{\Theta_{q,e}\}_{e=1}^{E} \gets \mathrm{Derive}(\mathbf{\Theta_{q}})$
        \hfill $\triangleright$ construct MoBi slices 
        
        \State Router score $\mathbf{S} \gets \mathrm{Router}(\mathbf{H}_{\mathrm{q}};\mathbf{\Theta_{r}})$
        \hfill $\triangleright$ token wise gating weights
        
        \State $\mathbf{Y}_{\mathrm{q}} \gets \mathbf{Y}_{\mathrm{FS}}
        \;+\; \sum_{e=1}^{E} \mathbf{G}_e \odot f_{e}(\mathbf{H}_{\mathrm{q}};\Theta_{q,e})$
        \hfill $\triangleright$ sum all slices' outputs
        
        \State $\mathcal{L}_{\mathrm{MoBi}}
        \gets \|\mathbf{Y}_{\mathrm{q}} - \mathbf{Y}_{\mathrm{fp}}\|_2^2
        + \lambda\,\mathrm{Reg}(\mathbf{S}, \mathbf{\Theta_{r}})$
        \hfill $\triangleright$ reconstruction loss and router regularization
        
        \State Update $\mathbf{\Theta_{q}}, \mathbf{\Theta_{r}}$
        \hfill $\triangleright$ joint update of quantization and router params
    \EndFor

    \State $\widehat{\mathbf{W}} \gets \mathrm{Quantize}(\mathbf{W};\mathbf{\Theta_{q}}, \mathbf{\Theta_{r}})$

    \State $\mathbf{H}_{\mathrm{fp}} \gets \mathbf{Y}_{\mathrm{fp}}$
    \State $\mathbf{H}_{\mathrm{q}} \gets f_{\mathrm{q}}(\mathbf{H}_{\mathrm{q}};\widehat{\mathbf{W}})$

\EndFor

\State \Return $\widehat{\mathcal{M}}$

\end{algorithmic}
\end{algorithm}
\vspace{-1.0em}

To ensure consistent refinement across slices, \texttt{MoBiSlice} iteratively refines the scaling factor so that each slice represents progressively finer increments. Let $(s_0,z_0)$ be the calibrated parameters of the first slice. After assigning $b_e$ bits to slice $e$, the next slice refines the resolution as $s_{e+1}=s_e/2^{b_e}$. Finally, while the first slice uses the calibrated zero point $z_0$, slices fix $z_e = 2^{b_e-1}$ for $e\ge1$, placing the midpoint code at the center of the integer range so that positive and negative residual corrections are represented symmetrically, which avoids systematic drift during slice accumulation. Together, the floor aligned mapping, scale refinement, and centered residual quantization make the slice decomposition compatible with truncation based precision control, enabling stable elastic reconstruction by activating a subset of slices.

\paragraph{Bias and Error Bounds for Bit Slice Activation.}
We now analyze truncation-based bit slice activation and show that activating residual slices performs a true refinement: it leaves the coarse quantization code unchanged and only adds a bounded zero-mean remainder term that recovers finer detail. First, we assume a two slice scenario, consisting of a shared base slice with $(b-k)$ bits and a single $k$ bit residual slice that can be activated or dropped.

Let $W$ be the original weight. The base slice produces an MSB code with parameters $(s_1,z_1)$, and the residual slice quantizes the residual $R$ using parameters $(s_2,z_2)$ derived from the base slice:
\begin{align}
\mathrm{MSB}_{b-k} &\;=\; \Big\lfloor \frac{W}{s_1}+z_1 \Big\rfloor, \\
\mathrm{LSB}_{k} &\;=\; \Big\lfloor \frac{R}{s_2}+z_2 \Big\rfloor,
\qquad
s_2=\frac{s_1}{2^{k}},
\qquad
z_2 = 2^{k-1}.
\end{align}
Using the centered dequantization, the two-slice reconstruction is:
\begin{align}
\label{eq:mobislice_two_expert_recon}
\widehat{W}
&= s_1\!\left(\mathrm{MSB}_{b-k}-z_1+0.5\right)
 \;+\; s_2\!\left(\mathrm{LSB}_{k}-z_2+0.5\right).
\end{align}

Although the reconstruction in Eq.~\eqref{eq:mobislice_two_expert_recon} is written for a base $(b-k)$ bit slice plus a $k$ bit residual slice, our goal is to characterize \emph{any} intermediate precision obtained by activating only a subset of the residual information. We therefore introduce a generic truncation level $p$ that drops $p$ least significant bits from the merged code. This captures all partial activation cases: $p=0$ corresponds to using all $k$ residual bits, and larger $p$ corresponds to deactivating progressively more of the finest bits. We truncate $p$ least significant bits of the merged integer code $\mathrm{INT}_8 = (\mathrm{MSB}_{b-k} \ll k) + \mathrm{LSB}_{k}$ using a right shift:
\begin{equation}
I \;=\; \mathrm{INT}_8 \gg p \;=\; \Big\lfloor \frac{\mathrm{INT}_8}{2^{p}} \Big\rfloor,
\qquad
\mathrm{INT}_8 \;=\; 2^{p} I + r,
\qquad
r \in \{0,1,\dots,2^p-1\}
\label{eq:shift_decompose}
\end{equation}

Substituting Eq. \eqref{eq:shift_decompose} into Eq. \eqref{eq:mobislice_two_expert_recon} and noting that $s_2=s_1/2^{k}$, we can rearrange terms to express $\widehat{W}$ as a coarse quantizer applied to the truncated code $I$, plus an additive truncation noise term determined by the dropped remainder $r$:
\begin{align}
\widehat{W}
&= s_2\!\left(\mathrm{INT}_8 - 2^{k}z_1 + 0.5\right)
\nonumber\\
&= s_2\!\left(2^{p}I + r - 2^{k}z_1 + 0.5\right)
\nonumber\\
&= \underbrace{(2^{p}s_2)}_{\displaystyle s'}\!\left(
I - \underbrace{2^{k-p}z_1}_{\displaystyle z'} + \underbrace{0.5}_{\displaystyle \mathrm{mid}}
\right)
\;+\;
\underbrace{s_2\!\left(r + 0.5 - 2^{p-1}\right)}_{\displaystyle E_p}.
\label{eq:shifted_recon_with_noise}
\end{align}

Therefore, truncating $p$ least significant bits is equivalent to quantizing with coarser parameters:
\begin{equation}
s' = 2^{p} s_2,
\qquad
z' = 2^{k-p} z_1,
\qquad
\mathrm{mid}=0.5,
\end{equation}
Eq.~\eqref{eq:shifted_recon_with_noise} makes the role of $p$ explicit: truncation does not change the coarse code $I$, and its only effect is an additive perturbation $E_p$ determined by the discarded remainder $r$. Thus, analyzing $E_p$ directly characterizes the impact of partially activating bit slices. Assuming the residue $r$ is approximately uniform over $\{0,1,\dots,2^{p}-1\}$, we obtain:
\begin{align}
\mathbb{E}[E_p]
&= s_2\!\left(\mathbb{E}[r] + 0.5 - 2^{p-1}\right)
= s_2\!\left(\frac{2^{p}-1}{2} + 0.5 - 2^{p-1}\right)
= 0,
\end{align}
This yields $\mathbb{E}[E_p]=0$, indicating that slice-based reconstruction does not introduce bias in expectation.
Moreover, since $r \in \{0,1,\dots,2^p-1\}$,
\begin{equation}
-\,\Big(2^{p-1}-0.5\Big) \;\le\; r + 0.5 - 2^{p-1} \;\le\; \Big(2^{p-1}-0.5\Big),
\end{equation}
which implies the error bound:
\begin{equation}
|E_p| \;\le\; s_2\Big(2^{p-1}-0.5\Big) \;<\; 2^{p-1}s_2
\quad\Longrightarrow\quad
\frac{|E_p|}{s_2 2^p} \;<\; \frac{1}{2}.
\label{eq:half_step_bound}
\end{equation}
Eq. \eqref{eq:half_step_bound} implies that the truncation error is strictly smaller than one half step of the coarser quantizer. Equivalently, the perturbation $E_p$ cannot move $\widehat{W}$ across a decision boundary of the base MSB quantizer, and therefore cannot flip any bit of the coarse code. As a result, activating residual slices performs a true residual refinement: it only fills in finer bit slices without altering the coarser representation, yielding an accurate and stable activation with controlled error. Therefore, this establishes that bit slice activation yields a guaranteed residual refinement: it adds finer information while preserving the coarse code, and its only effect is an additive truncation noise term that is bounded and zero-mean.
\paragraph{Prefix Selection Is Naturally Learned.}
The formulation in Eq.~\ref{eq:recon} requires selecting a prefix of the ordered slices. In practice, we do not enforce prefix selection, allowing token-aware precision routing to use $E$ independent binary gates and enabling run-time precision adjustment through a single global threshold, as in Eq.~(10). In contrast, parameterizing the routing decision as a one-hot selection among $E$ prefix lengths complicates such threshold-based adjustment. Nonetheless, we find that an explicit prefix constraint is unnecessary, as prefix selection emerges naturally during optimization. Earlier slices tend to receive stronger optimization signals, encouraging the router to preferentially select them when minimizing the reconstruction loss.

Consider four 2-bit slices $[\mathbf{W}_1,\mathbf{W}_2,\mathbf{W}_3,\mathbf{W}_4]$, ordered from the most significant to the least significant bit-slice. Given an input token $\mathbf{X}$, a shared scale $\mathbf{\Theta}_q$, router parameters $\mathbf{\Theta}_r$, and an upstream gradient $g$, the routing and reconstruction process can be written as
\begin{equation}
    \mathbf{X}\mathbf{\Theta}_r
    \;\longrightarrow\;
    \left(
        2^6\mathbf{W}_1
        + 2^4\mathbf{W}_2
        + 2^2\mathbf{W}_3
        + 2^0\mathbf{W}_4
    \right)
    \mathbf{X}\mathbf{\Theta}_q,
\end{equation}
with branch-wise gradient contributions proportional to
\begin{equation}
    \left[
        2^6\mathbf{W}_1 g\mathbf{X},\;
        2^4\mathbf{W}_2 g\mathbf{X},\;
        2^2\mathbf{W}_3 g\mathbf{X},\;
        2^0\mathbf{W}_4 g\mathbf{X}
    \right].
\end{equation}
Thus, earlier slices have a larger influence on reconstruction-error reduction. At the beginning of training, all slices are active. As the regularization term gradually encourages fewer active slices, the router retains the slices that reduce the L2 loss in Eq.~\ref{equ:train_objective} the most. For example, if a token is routed to 4-bit precision using two slices, selecting $\mathbf{W}_1+\mathbf{W}_2$ preserves the most significant bit-slices, whereas $\mathbf{W}_3$ and $\mathbf{W}_4$ contribute exponentially less to the reconstruction. Therefore, prefix selection naturally yields the smallest reconstruction error, making non-prefix combinations suboptimal. Consequently, \texttt{MoBiRoute} learns prefix selection without an explicit constraint. Empirically, after convergence, all routed token--bit assignments satisfy the prefix property, selecting the first $k$ slices for $2k$-bit routing.
\section{Experiment Details}
\label{app:exp_detail}
 
\subsection{MoBiQuant Training Setup}
\label{app:setup}
We use weight-only quantization in all reported runs in the main paper, with \texttt{abits}=16 and \texttt{group\_size}=128, while the base bit slice uses \texttt{wbits}=2. Weight-activation quantization results and their experimental setup are reported in Appendix~\ref{app:activation}.
Slice decomposition is enabled by \texttt{slice\_bits\_list}, and our default configuration uses four bit slices with \texttt{slice\_bits\_list = 2\;2\;2\;2}. Training proceeds for \texttt{epochs}=20 and \texttt{nsamples}=128, with \texttt{batch\_size}=1 for all models. The total number of router annealing steps is used by the router temperature schedule and calculated as follows:
\[
\texttt{global\_steps} = \Big(\frac{\texttt{nsamples}}{\texttt{batch\_size}}\Big)\cdot \texttt{epochs},
\]
For each layer, we optimize three parameter groups with AdamW: learnable weight clipping parameters (LWC), learnable equivalent transformation parameters (LET), and MoBiQuant parameters. Group specific learning rates are denoted as follows: \texttt{lwc\_lr}, \texttt{let\_lr}, and \texttt{mobi\_lr}. We typically use \texttt{lwc\_lr} in the range \texttt{1e\textminus{}3} to \texttt{1e\textminus{}2}, and \texttt{mobi\_lr} in the range \texttt{5e\textminus{}6} to \texttt{4e\textminus{}5}, depending on model size.

\subsection{Flexible Inference and Effective Bit Control}
\label{app:flexible}
A core feature of \texttt{MoBiQuant} is flexible inference, where the effective precision is controlled by activating a subset of residual bit slices. At evaluation time, this is exposed by \texttt{target\_activation\_ratio\_eval} or \texttt{target\_bit\_eval}. When \texttt{target\_bit\_eval} is used, we convert it into a desired routing ratio based on the base MSB bits and the sum of residual slice bits:
\[
\rho \;=\; \frac{\texttt{target\_bit\_eval} - b_{\mathrm{msb}}}{\sum_{e\ge 1} b_e},
\qquad b_{\mathrm{msb}} = \texttt{slice\_bits\_list}[0].
\]
Hard binary gating is applied in evaluation using router scores and a threshold selected to satisfy the target ratio.

\paragraph{Layer-wise threshold calibration.}
To make the realized routing ratio stable across layers and modules, we calibrate a layer specific threshold from calibration samples whenever \texttt{target\_bit\_eval} requests residual activation. This calibration collects router scores over the calibration set and sets each layer's threshold to the appropriate quantile so that approximately a fraction $\rho$ of routed experts are active. We set the calibration sample to be the same as \texttt{nsamples}.
\subsection{Baseline Results}
\label{app:baseline_results}

The baseline results reported in our experiments are either taken from the corresponding published papers or obtained by running their official open-source implementations. All reproduced experiments are conducted on NVIDIA A100 GPUs. Specifically, for OmniQuant, SpinQuant, AWQ, GPTQ, SmoothQuant, QUIP\#, and QTIP, we run their open-source code to obtain WikiText2 PPL results on the LLaMA-3.2-1B, LLaMA-3.2-3B, and LLaMA-3-8B model series under a context length of 2048. In particular, for QUIP\#, we reproduce and report its non-finetuned variant to align its computational cost with the other post-training quantization baselines. These reproduced results are used to ensure a consistent evaluation setting with our method. For kernel measurement, to ensure a consistent evaluation environment with our customized kernel, we uniformly evaluate the kernels of QUIP\#, QTIP, AnyPrecisionLLM, and AnyBCQ on an NVIDIA A100 GPU with CUDA 12.9.
\section{Ablation Studies}
\label{app:ablation}
\subsection{Ablation of Calibration Dataset}
\label{app:downstream}
\begin{table*}[htbp]
\captionsetup{justification=justified}
\caption{\textbf{Ablation study on different calibration datasets for LLaMA3.2-1B-3bit.}}
\vspace{-5pt}
\label{tab:calib_ablation_llama32_1b}
\centering
\scriptsize
\setlength{\tabcolsep}{3pt}
\renewcommand{\arraystretch}{1.1}

\begin{tabular}{c|c|ccc|cccccccc}
\toprule

\multirow{2}{*}{\textbf{Dataset}} &
\multirow{2}{*}{\textbf{Method}} &
\multicolumn{3}{c|}{\textbf{Perplexity (PPL $\downarrow$)}} &
\multicolumn{8}{c}{\textbf{Downstream Tasks (Acc $\uparrow$)}} \\

& & 
Wiki & C4 & PTB &
ARC E & ARC E(n) &
ARC C & ARC C(n) &
BoolQ &
Hella & Hella(n) &
Wino \\
\midrule

\multirow{2}{*}{Wiki}
& Omni. 
& 21.597 & 32.860 & 37.467
& 0.510 & 0.472
& 0.245 & 0.284
& 0.593
& 0.378 & 0.488
& \textbf{0.551} \\

& MoBiQ &  \textbf{18.947} & \textbf{27.924} & \textbf{34.915}
& \textbf{0.521} & \textbf{0.481}
& \textbf{0.261} & \textbf{0.294}
& \textbf{0.613}
& \textbf{0.382} & \textbf{0.497}
& 0.533 \\
\midrule

\multirow{2}{*}{C4}
& Omni. 
& \textbf{22.611} & \textbf{33.198} & \textbf{38.096}
& 0.503 & \textbf{0.470}
& 0.245 & 0.282
& 0.583
& \textbf{0.378} & \textbf{0.488}
& \textbf{0.548} \\

& MoBiQ &  26.360 & 33.382 & 43.888
& \textbf{0.522} & 0.452
& \textbf{0.282} & \textbf{0.293}
& \textbf{0.619}
& 0.369 & 0.477
& 0.538 \\
\midrule

\multirow{2}{*}{PTB}
& Omni. 
& 22.689 & 33.519 & 35.353
& 0.503 & 0.471
& 0.241 & 0.281
& 0.586
& 0.376 & 0.488
& \textbf{0.556} \\

& MoBiQ  &\textbf{16.909} & \textbf{24.849} & \textbf{29.268}
& \textbf{0.588} & \textbf{0.532}
& \textbf{0.284} & \textbf{0.312}
& \textbf{0.589}
& \textbf{0.402} & \textbf{0.526}
& \textbf{0.556} \\
\midrule

\multirow{2}{*}{Mix}
& Omni. 
& 22.335 & 33.327 & 36.763
& 0.502 & 0.474
& 0.246 & 0.277
& 0.599
& 0.374 & 0.488
& \textbf{0.553} \\

& MoBiQ 
& \textbf{21.070} & \textbf{30.072} & \textbf{33.646}
& \textbf{0.534} & \textbf{0.502}
& \textbf{0.258} & \textbf{0.300}
& \textbf{0.612}
& \textbf{0.382} & \textbf{0.500}
& 0.530 \\

\bottomrule
\end{tabular}
\end{table*}

We study the impact of different calibration datasets in Tab.~\ref{tab:calib_ablation_llama32_1b}. 
\texttt{MoBiQuant} remains stable across calibration choices and consistently achieves lower perplexity than OmniQuant on most calibration--evaluation pairs. 
In particular, WikiText2 and PTB calibration provide the strongest improvements across WikiText2, C4, and PTB, while the mixed calibration setting maintains similar average bits with favorable cross-domain generalization.

On zero-shot tasks, \texttt{MoBiQuant} generally matches or outperforms OmniQuant across ARC Easy, ARC Challenge, BoolQ, HellaSwag, and WinoGrande~\cite{clark2018think, clark2019boolq, zellers2019hellaswag}. 
The most consistent gains are again observed with PTB and WikiText2 calibration, indicating that \texttt{MoBiQuant} is robust to calibration set selection across both language modeling and downstream reasoning tasks.
\subsection{Ablation of Scheduling Methods for Router Regularization}
\label{app:schedule}
\begin{figure*}[h]
    \centering
    \includegraphics[width=\linewidth]{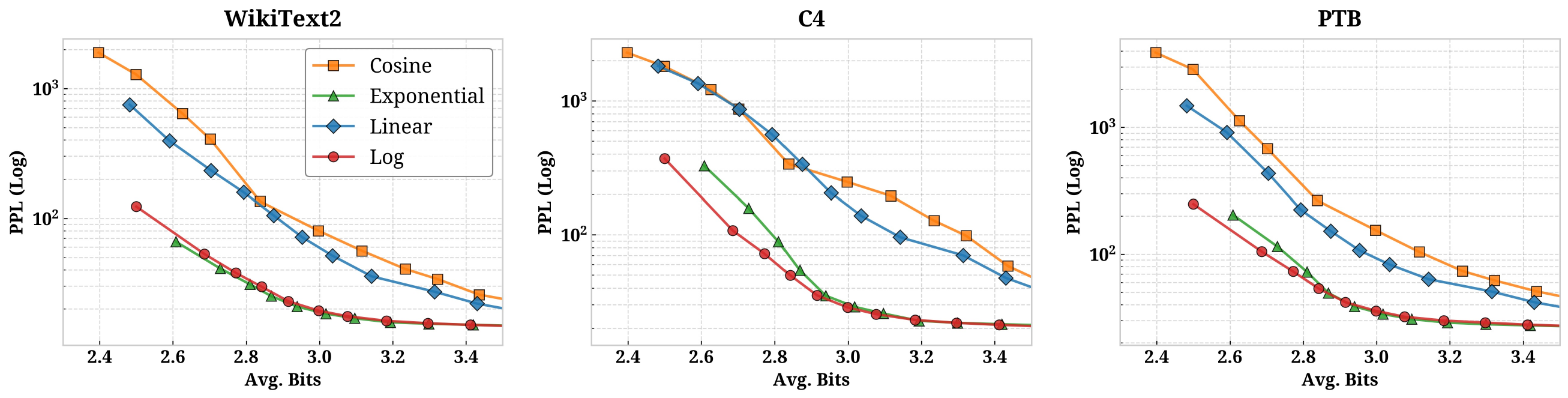}
    \caption{Comparison of PPL ($\downarrow$) sweep between different scheduling methods across WikiText2, C4, PTB datasets.}
    \label{fig:schedule}
\end{figure*}

We conduct an ablation study to identify the most effective scheduling strategy for router regularization. We compare four schedules, Linear, Cosine, Exponential, and Logarithmic, and measure their impact on model perplexity (PPL) on WikiText2, C4, and PTB. As shown in Fig.~\ref{fig:schedule}, performance differences are most pronounced in the low-precision regime of 2.5 to 3.0 average effective bits. \textbf{In this range, both Logarithmic and Exponential scheduling consistently outperform the Linear and Cosine alternatives.} The relative ordering of the two best schedules is dataset dependent: Exponential offers a slight advantage on WikiText2, while Logarithmic performs better on C4 and PTB within the same bit-width range.

We adopt a logarithmic schedule for regularization, as it naturally aligns with the logarithmic temperature annealing used in router gating and converges smoothly to the target precision during inference. 
This synchronization between routing sharpness and precision allocation stabilizes training and enables smoother adaptation to resource constraints.

\subsection{Ablation of Different Training Target Bit}
\label{app:target}
We train LLaMA3.2-1B with five target bit budgets ($2.5$, $3.0$, $3.5$, $4.0$, and $5.0$) to study how the training target affects router generalization across precisions. 
For each configuration, we sweep inference-time budgets and evaluate PPL on WikiText2, C4, and PTB. 

\begin{figure*}[ht]
    \centering
    \includegraphics[width=\linewidth]{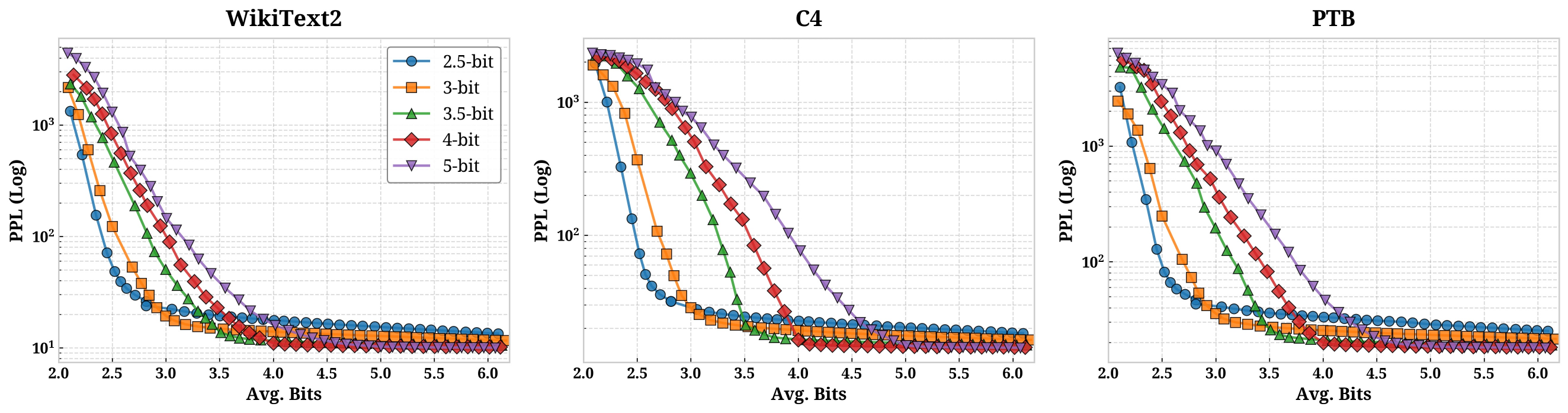}
    \caption{Comparison of PPL ($\downarrow$) sweep between different target bit-widths for training across WikiText2, C4, PTB datasets.}
    \label{fig:target}
\end{figure*}

As shown in Fig.~\ref{fig:target}, consistent trends are observed across datasets, with the largest differences appearing near the transition region around $3.0$ average bits. 
Among all settings, the $3.0$-bit target achieves the best overall trade-off. 
Lower targets (e.g., $2.5$-bit) improve low-bit performance but degrade higher-bit PPL, while higher targets ($3.5$--$5.0$-bit) significantly worsen sub-$3.0$-bit performance with only marginal gains at higher precisions.
Elastic inference requires the router to accommodate a changing target budget, and the target precision scheduling objective in Eq.~\ref{eq:reg} is designed to encourage exploration of diverse token slice allocations before converging. However, the training target still sets the operating point around which the router learns stable binary decisions: a low target biases the router toward early and persistent slice pruning, while a high target biases it toward maintaining many active slices. The $3.0$ bit target sits near a balanced regime and improves calibration of routing decisions across a broader range of average bits. This motivates us to adopt $3.0$ bits as the default training target throughout all experiments.

\section{Additional Experiments}
\label{app:exp}
\subsection{ Outlier migration in Non-Gradient-Based PTQ}
\label{app:AWQ}
To further validate the generality of our observation, we analyze the behavior of non-gradient-based PTQ methods under calibration--inference mismatch. 
We take AWQ \citep{lin2024awq} as a representative baseline and evaluate its generalization performance across different bit-widths.
\begin{table}[ht]
\centering
\caption{Generalization gap (PPL) of AWQ on LLaMA2-7B.}
\label{tab:awq_gap}
\begin{tabular}{c|cc}
\hline
\textbf{Calibration Bits} & \textbf{Infer @ 3-bit} & \textbf{Infer @ 4-bit} \\
\hline
3-bit & 8.22 & 7.85 \\
4-bit & 8.45 & 7.36 \\
\hline
\end{tabular}
\end{table}
As shown in Table~\ref{tab:awq_gap}, AWQ exhibits noticeable degradation when the inference precision differs from the calibration setting, indicating limited generalization across bit-widths. 
In addition, we observe that the top outlier tokens overlap by only 41\% between 3-bit and 4-bit settings, suggesting that token sensitivity shifts significantly across precisions. 
This behavior is consistent with the \textit{outlier migration} phenomenon described in the main text.

\subsection{Outlier migration on Mistral-7B}
\label{app:mistral}
We further verify that the outlier migration phenomenon generalizes beyond the LLaMA family. 
Specifically, we observe similar behavior on Mistral-7B. 
For example, in a middle layer of Mistral-7B, the overlap of top token outliers between 3-bit and 4-bit is only 16\%, indicating significant precision-dependent variation in token sensitivity. 
This migration leads to poor generalization for existing PTQ methods. 
As shown in Table~\ref{tab:mistral_ppl}, when the inference bit-width differs from the calibration setting, performance degrades substantially. 
In contrast, \texttt{MoBiQuant} effectively mitigates this issue.

\begin{table}[ht]
\centering
\caption{Mistral-7B performance gap (PPL) under calibration--inference mismatch.}
\label{tab:mistral_ppl}
\begin{tabular}{lcc}
\hline
\textbf{Method} & \textbf{Calib 3 $\rightarrow$ Infer 4} & \textbf{Calib 4 $\rightarrow$ Infer 3} \\
\hline
OmniQuant & 5.61 & 23.19 \\
MoBiQuant & \textbf{5.42} & \textbf{19.38} \\
\hline
\end{tabular}
\end{table}

\subsection{Compatibility with rotation-based methods}
\label{app:quarot}
We further investigate whether our method can be combined with rotation-based PTQ techniques. 
We consider two representative methods: (1) the non-gradient-based QuaRot \citep{ashkboos2024quarot}, and (2) the gradient-based DuQuant \citep{lin2024duquant}.

\begin{table}[ht]
\centering
\caption{Generalization gap (PPL) of QuaRot on LLaMA2-7B.}
\label{tab:quarot}
\begin{tabular}{c|cc}
\hline
\textbf{Method} & \textbf{Calib 4 $\rightarrow$ Infer 4} & \textbf{Calib 4 $\rightarrow$ Infer 3} \\
\hline
OmniQ & 5.74 & 32.80 \\
OmniQ + QuaRot & 5.65 & 27.55 \\
MoBiQuant + QuaRot & \textbf{5.65} & \textbf{22.24} \\
\hline
\end{tabular}
\end{table}
Table~\ref{tab:quarot} shows that while QuaRot improves static performance, the degradation under bit-width mismatch remains significant. 
In contrast, combining \texttt{MoBiQuant} with QuaRot substantially improves generalization, demonstrating that our approach is complementary to rotation-based techniques.
\begin{table}[htb]
\centering
\caption{Generalization performance of DuQuant and \texttt{MoBiQuant} on unseen W-A configurations.}
\label{tab:duquant}
\begin{tabular}{c|ccc|ccc}
\hline
 & \multicolumn{3}{c|}{\textbf{LLaMA2-7B}} & \multicolumn{3}{c}{\textbf{LLaMA3-8B}} \\
\textbf{W-A} & W3A4 & W4A4 & W5A4 & W3A4 & W4A4 & W5A4 \\
\hline
DuQuant & 6.77 & 6.51 & 6.49 & 17.12 & 15.31 & 12.12 \\
MoBiQuant + DuQuant & 6.81 & \textbf{6.40} & \textbf{6.30} & \textbf{13.71} & \textbf{12.25} & \textbf{11.69} \\
\hline
\end{tabular}
\end{table}
For DuQuant, we calibrate the model at W3A4 and evaluate on unseen weight--activation configurations. 
As shown in Table~\ref{tab:duquant}, \texttt{MoBiQuant} consistently improves generalization across all settings, with significant gains on LLaMA3-8B.
Overall, these results demonstrate that (1) the outlier migration phenomenon is not limited to a specific PTQ method, and (2) \texttt{MoBiQuant} effectively improves generalization when combined with existing PTQ techniques.

\subsection{Any-precision performance with Activation Quantization}
\label{app:activation}

\begin{figure}[htb]
  \centering
  \includegraphics[width=0.44\columnwidth]{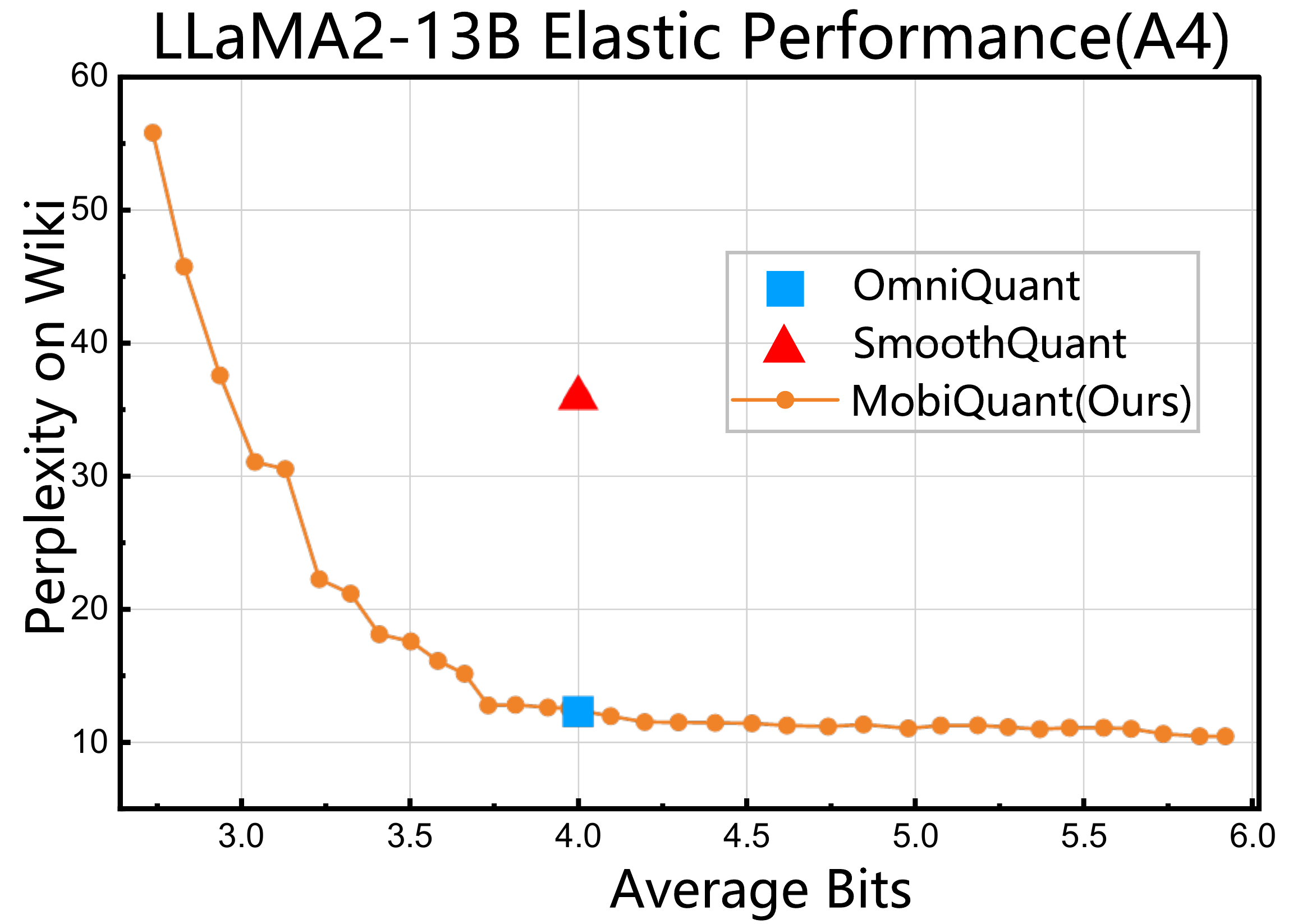}
  \caption{PPL ($\downarrow$) tradeoff under 4 bit activation quantization on LLaMA2-13B. We compare with SmoothQuant~\cite{xiao2023smoothquant} and OmniQuant~\cite{shao2023omniquant} W4A4 baselines.}
  \label{fig:actquant}
\end{figure}

The elasticity of \texttt{MoBiQuant} also extends to activation quantization, as shown in Fig.~\ref{fig:actquant}, where we report results on LLaMA2-13B. We keep the same slice decomposition as the weight-only setting and the same calibration hyperparameters. We also keep the same regularization parameter and learning rate. 
A key practical detail in this setting is the interaction between learnable equivalent transforms (LET) from OmniQuant~\cite{shao2023omniquant} and router training. LET preserves the main linear path by transforming the input activation and compensating the linear weights so that the layer output remains equivalent. For a linear layer,

\begin{align}
\mathbf{Y}=\mathbf{X}\mathbf{W}+\mathbf{B},
\qquad
\tilde{\mathbf{X}}=(\mathbf{X}-\boldsymbol{\delta})\oslash\mathbf{s},
\qquad
\tilde{\mathbf{W}}=\mathbf{s} \odot \mathbf{W},
\qquad
\tilde{\mathbf{B}}=\mathbf{B}+\boldsymbol{\delta}\mathbf{W},
\end{align}

where $\mathbf{X}\in\mathbb{R}^{T\times d}$ denote the input activations with $T$ tokens and $d$ input channels; $\mathbf{W}\in\mathbb{R}^{d\times m}$ the weight matrix; and $\mathbf{B}\in\mathbb{R}^{m}$ the bias. LET introduces a learnable per-channel shift $\boldsymbol{\delta}\in\mathbb{R}^{d}$ and scale $\mathbf{s}\in\mathbb{R}^{d}$, producing the transformed input $\tilde{\mathbf{X}}$ and the compensated parameters $\tilde{\mathbf{W}}$, $\tilde{\mathbf{B}}$ so that the layer output is preserved.
However, the router is a separate linear branch whose weights are not re-parameterized by LET, so feeding $\tilde{\mathbf{X}}$ directly changes the routing function and can distort router scores. This is important because LET parameters evolve during calibration, so the router would otherwise see a moving input distribution, which can drive logits into saturation under sharp gating schedules and small calibration budgets, especially when the router input is quantized. For stability, we undo LET for the router input by reconstructing the activation in the original space before routing as follows:
\begin{align}
\mathbf{X}_{r}=\tilde{\mathbf{X}}\odot\mathbf{s}+\boldsymbol{\delta},
\end{align}
This keeps the router operating in a fixed activation space while LET continues to re-parameterize the main linear path, improving routing stability.

\subsection{Calibration cost comparison}
\label{app:calibration_cost}

We report in Tab.~\ref{tab:calibration_cost} the calibration time and computational cost required to obtain the $\{2,3,4\}$-bit LLaMA2-7B models. \texttt{MoBiQuant} requires only a single calibration run with 128 samples, comparable to calibrating one static PTQ model, to support multiple bit-widths. In contrast, OmniQuant requires three independent calibration runs, substantially increasing both calibration time and total calibration-data usage.
\texttt{MoBiQuant} also incurs lower calibration-memory overhead than AnyBCQ. AnyBCQ maintains multiple groups of learnable scaling factors associated with different bit-planes, whereas \texttt{MoBiQuant} shares a single scaling-factor parameterization across slices. This design eliminates precision-specific scaling parameters and their associated optimizer states, reducing the peak calibration memory from 19 GB to 17 GB.
Overall, all three methods can be calibrated on a single NVIDIA RTX 4090 GPU with 24 GB of memory, demonstrating that the offline calibration cost remains practical for real-world deployment.

\begin{table}[htb]
    \centering
    \small
    \caption{Calibration time and computational cost for obtaining the $\{2,3,4\}$-bit LLaMA2-7B models on a single NVIDIA RTX 4090 GPU.}
    \label{tab:calibration_cost}
    \begin{tabular}{c|cccc}
        \toprule
        \textbf{Method} &
        \textbf{Time} &
        \textbf{Data Usage} &
        \textbf{Peak Memory} &
        \textbf{Calibration Runs} \\
        \midrule
        OmniQuant    & 7.6 h          & 384          & 13 GB & 3 \\
        AnyBCQ       & 6.3 h          & 512          & 19 GB & 1 \\
        MoBiQuant    & \textbf{4.5 h} & \textbf{128} & 17 GB & 1 \\
        \bottomrule
    \end{tabular}
\end{table}

\subsection{Additional Comparison with Static PTQ Methods} 
\label{app:staticptq}

\paragraph{Vector Quantization Methods.}
We further compare \texttt{MoBiQuant} with the recent VQ-based methods QuIP\#~\citep{tseng2024quip} and QTIP~\citep{tseng2024qtip}. Although these methods achieve strong bit--accuracy trade-offs, their decoding throughput is limited by centroid-table lookup overhead. As shown in Tab.~\ref{tab:vqcomparison}, 2-bit QuIP\# and QTIP achieve 191 and 218 tokens/sec.\ on LLaMA2-7B, with perplexities of 12.30 and 5.86, respectively. In comparison, 4-bit \texttt{MoBiQuant} achieves a lower perplexity of 5.82 while increasing throughput to 256 tokens/sec. Overall, \texttt{MoBiQuant} provides a favorable accuracy--throughput trade-off while retaining run-time precision adjustment.

\begin{table}[htb]
\vspace{-7pt}
\centering
\footnotesize
\caption{Comparison with the static VQ methods QuIP\# and QTIP. 
\texttt{MoBiQuant} provides competitive perplexity with higher decoding throughput while supporting run-time precision adjustment.}
\label{tab:vqcomparison}
\begin{tabular}{cc|cccc|ccc}
\toprule
\multirow{2}{*}{\textbf{Model}} &
\multirow{2}{*}{\textbf{Bits}} &
\multicolumn{4}{c|}{\textbf{WikiText2, PPL $\downarrow$}} &
\multicolumn{3}{c}{\textbf{Throughput, tokens/sec. $\uparrow$}} \\
\cmidrule(lr){3-6}
\cmidrule(lr){7-9}
& & FP16 & QuIP\# & QTIP & MoBiQ
  & QuIP\# & QTIP & MoBiQ \\
\midrule

\multirow{3}{*}{LLaMA2-7B}
& 2 & 5.47 & 12.30 & \textbf{5.86} & 10.91
    & 191 & 218 & \textbf{404} \\
& 3 & 5.47 & 6.19 & \textbf{5.28} & 6.07
    & 177 & 196 & \textbf{311} \\
& 4 & 5.47 & 5.66 & \textbf{5.17} & 5.82
    & 155 & 189 & \textbf{256} \\
\midrule

\multirow{3}{*}{LLaMA3-8B}
& 2 & 6.12 & 15.20 & \textbf{7.25} & 58.12
    & 157 & 180 & \textbf{352} \\
& 3 & 6.12 & 8.27 & \textbf{7.05} & 7.97
    & 142 & 158 & \textbf{257} \\
& 4 & 6.12 & 6.64 & \textbf{6.06} & 7.31
    & 126 & 154 & \textbf{209} \\

\bottomrule
\end{tabular}
\vspace{-9pt}
\end{table}

\paragraph{Downstream Evaluation.}
We further compare elastic \texttt{MoBiQuant} with static PTQ baselines on downstream zero-shot tasks. Table~\ref{tab:staticcomparisonzeroshot} reports the average accuracy over six commonsense-reasoning tasks. Prior methods use fixed 4-bit precision, whereas the average precision of \texttt{MoBiQuant} is restricted to 3.9--4.0 bits for a fair comparison. \texttt{MoBiQuant} remains competitive with the strongest static PTQ methods and outperforms several widely used baselines across both LLaMA2 and LLaMA3 models. Table~\ref{tab:gsm8k} further reports GSM8K performance, where the elastic \texttt{MoBiQuant} outperforms static 4-bit OmniQuant. These results demonstrate that \texttt{MoBiQuant} preserves run-time precision flexibility while maintaining strong downstream-task performance.

\begin{table*}[htbp]
\captionsetup{justification=justified} 
\caption{Zero-shot comparison with static scalar quantization (SQ) methods. We report average accuracy over six zero-shot commonsense reasoning tasks.MoBiQuant matches or surpasses static PTQ baselines. ``-'' denotes that no direct results are reported in the published papers.}
\label{tab:staticcomparisonzeroshot}

\vspace{-5pt}
\centering
\footnotesize
\setlength{\tabcolsep}{2pt} 
\begin{tabular}{c|c|c|c|c|c|c|c}
\toprule
\multirow{3}{*}{Method} &
\multirow{3}{*}{Avg. Bits} & 
\multirow{3}{*}{Type} &
LLaMA-2-7B &  LLaMA-2-13B & LLaMA-3.2-1B & LLaMA-3.2-3B & LLaMA-3-8B   \\
 & & & 0-shot  & 0-shot  & 0-shot  & 0-shot & 0-shot  \\
\scriptsize & & & Avg.($\uparrow$)  & Avg.($\uparrow$)  & Avg.($\uparrow$)  & Avg.($\uparrow$)  & Avg.($\uparrow$)   \\
\midrule
Floating-point & 16 & - & 69.68  & 73.19  & 59.42  & 66.71  & 74.19     \\ \midrule
RTN & \multirow{6}{*}{4} & \multirow{6}{*}{Static SQ} & 64.72  & 64.19    & -  & -  & 70.40    \\ 
SmoothQuant & & & 63.87  & 67.46   & - & - & 67.47   \\
AWQ  & &  &\textbf{69.40}  & 72.64   & - & -  & \textbf{72.64}    \\
GPTQ & & & 65.54  & 70.22   & -  & - & 70.21   \\
SpinQuant & & & 68.00  & \textbf{72.68}  & - & -   & 72.62  \\
OmniQuant & & & 68.41  & 71.68   & \textbf{56.99}  & \textbf{63.8}  & 71.66  \\
\midrule
MoBiQuant & 3.9-4.0 &AnyPrec. &67.22  &71.54   & 55.49  & 63.2  &71.84  \\
\bottomrule
\end{tabular}
\end{table*}
\begin{table}[htbp]
\centering
\caption{GSM8K performance of LLaMA3.2-1B under 4-bit setting.}
\label{tab:gsm8k}
\begin{tabular}{lcc}
\hline
\textbf{Method} & \textbf{Flexible-Extract} & \textbf{Strict-Match} \\
\hline
FP16 & 46.47\% & 42.15\% \\
OmniQuant-4bit & 30.86\% & 30.86\% \\
Ours (Elastic) & \textbf{32.83\%} & \textbf{32.45\%} \\
\hline
\end{tabular}
\end{table}

\subsection{Memory Savings under Multi-Precision Deployment.}\label{app:MemSavings}
\begin{figure}[htbp]
    \centering
\includegraphics[width=0.4\linewidth]{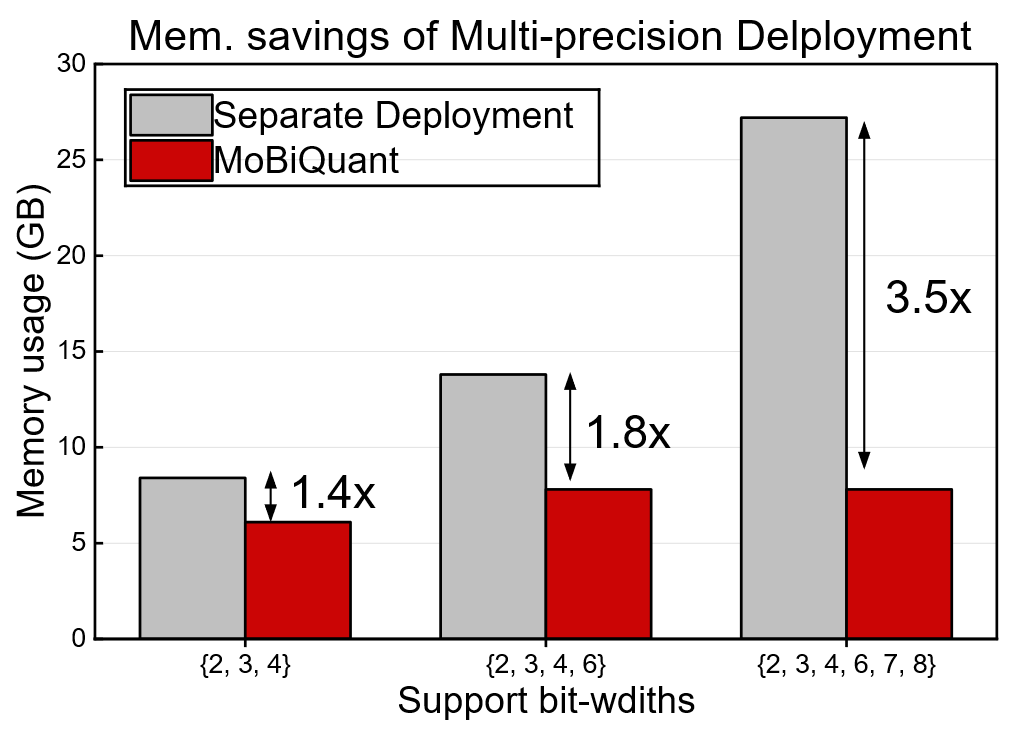}
    \caption{\textbf{Memory savings under multi-precision deployment.}
    Conventional deployment stores a separate quantized model for each supported precision, whereas \texttt{MoBiQuant} supports multiple precisions within a single composable representation, reducing the total model memory footprint by up to $3.5\times$.}
    \label{fig:memory_savings}
    \vspace{-5pt}
\end{figure}

We further evaluate the memory efficiency of \texttt{MoBiQuant} under multi-precision deployment. Conventional approaches require separately storing a quantized model for each supported precision, causing the total memory footprint to grow with the number of precision options. In contrast, \texttt{MoBiQuant} represents multiple precisions within a single model by composing different prefixes of the shared bit-slices. It therefore supports run-time precision adjustment without storing separate weight copies. As shown in Fig.~\ref{fig:memory_savings}, this many-in-one representation reduces the total model memory footprint by up to $3.5\times$ compared with separate multi-precision deployment.
\section{Limitations}
\label{app:limitations}
While \texttt{MoBiQuant} demonstrates strong accuracy--efficiency trade-offs across multiple model families and bit budgets, several limitations remain. First, our experiments are primarily conducted on small- to medium-scale LLMs (up to 32B parameters). We have not yet evaluated the method on ultra-large models such as 70B-scale LLMs due to the substantial computational and memory requirements of training and benchmarking elastic multi-precision quantization systems at that scale. We leave large-scale evaluation as future work.
Second, although we implement optimized custom kernels for our method, the current system has not yet been fully integrated with mature production inference frameworks such as vLLM or FlashInfer. 
Finally, the behavior of token-wise adaptive precision under large-scale cloud serving workloads with heterogeneous batching strategies remains an important direction for future investigation.


\newpage
\section*{NeurIPS Paper Checklist}

\begin{enumerate}

\item {\bf Claims}
    \item[] Question: Do the main claims made in the abstract and introduction accurately reflect the paper's contributions and scope?
    \item[] Answer: \answerYes{} 
    \item[] Justification: Abstract, section \ref{sec:Introduction}

\item {\bf Limitations}
    \item[] Question: Does the paper discuss the limitations of the work performed by the authors?
    \item[] Answer: \answerYes{} 
    \item[] Justification: Appendix \ref{app:limitations}

\item {\bf Theory assumptions and proofs}
    \item[] Question: For each theoretical result, does the paper provide the full set of assumptions and a complete (and correct) proof?
    \item[] Answer: \answerYes{} 
    \item[] Justification: Section \ref{sec:Motivation}, \ref{Section5}

    \item {\bf Experimental result reproducibility}
    \item[] Question: Does the paper fully disclose all the information needed to reproduce the main experimental results of the paper to the extent that it affects the main claims and/or conclusions of the paper (regardless of whether the code and data are provided or not)?
    \item[] Answer: \answerYes{} 
    \item[] Justification: Appendix \ref{app:exp_detail}

\item {\bf Open access to data and code}
    \item[] Question: Does the paper provide open access to the data and code, with sufficient instructions to faithfully reproduce the main experimental results, as described in supplemental material?
    \item[] Answer: \answerYes{} 
    \item[] Justification: The paper provides a code link in the abstract, along with sufficient instructions to faithfully reproduce the main experimental results.

\item {\bf Experimental setting/details}
    \item[] Question: Does the paper specify all the training and test details (e.g., data splits, hyperparameters, how they were chosen, type of optimizer) necessary to understand the results?
    \item[] Answer: \answerYes{} 
    \item[] Justification: Section \ref{Section5},  Appendix \ref{app:exp_detail}

\item {\bf Experiment statistical significance}
    \item[] Question: Does the paper report error bars suitably and correctly defined or other appropriate information about the statistical significance of the experiments?
    \item[] Answer: \answerYes{} 
    \item[] Justification: Section \ref{Section5},  Appendix \ref{app:exp_detail}, \ref{app:exp}

\item {\bf Experiments compute resources}
    \item[] Question: For each experiment, does the paper provide sufficient information on the computer resources (type of compute workers, memory, time of execution) needed to reproduce the experiments?
    \item[] Answer: \answerYes{} 
    \item[] Justification: Section \ref{Section5},  Appendix \ref{app:exp_detail}
    
\item {\bf Code of ethics}
    \item[] Question: Does the research conducted in the paper conform, in every respect, with the NeurIPS Code of Ethics \url{https://neurips.cc/public/EthicsGuidelines}?
    \item[] Answer: \answerYes{}{} 
    \item[] Justification: 
The authors affirm that the research presented in this paper conforms to the NeurIPS Code of Ethics.


\item {\bf Broader impacts}
    \item[] Question: Does the paper discuss both potential positive societal impacts and negative societal impacts of the work performed?
    \item[] Answer: \answerNo{} 
    \item[] Justification: The proposed method is a systems and efficiency optimization technique for LLM quantization and deployment, and we do not identify any direct potential malicious or unintended uses beyond those already associated with general-purpose language models. Therefore, the paper does not include a dedicated societal impact discussion.
    
\item {\bf Safeguards}
    \item[] Question: Does the paper describe safeguards that have been put in place for responsible release of data or models that have a high risk for misuse (e.g., pre-trained language models, image generators, or scraped datasets)?
    \item[] Answer: \answerNA{} 
    \item[] Justification: The paper does not introduce new datasets or foundation models with elevated misuse risks, and therefore does not include additional safeguard mechanisms for release. All datasets and pre-trained models used in this work are publicly available open-source resources released by their original providers.

\item {\bf Licenses for existing assets}
    \item[] Question: Are the creators or original owners of assets (e.g., code, data, models), used in the paper, properly credited and are the license and terms of use explicitly mentioned and properly respected?
    \item[] Answer: \answerYes{} 
    \item[] Justification: All datasets, codebases, and pre-trained models used in this work are properly credited through citations to their original sources. We also follow the corresponding licenses and terms of use provided by the original authors and organizations.

\item {\bf New assets}
    \item[] Question: Are new assets introduced in the paper well documented and is the documentation provided alongside the assets?
    \item[] Answer: \answerNA{} 
    \item[] Justification: This work does not introduce new public datasets, models, or other released assets requiring additional documentation.

\item {\bf Crowdsourcing and research with human subjects}
    \item[] Question: For crowdsourcing experiments and research with human subjects, does the paper include the full text of instructions given to participants and screenshots, if applicable, as well as details about compensation (if any)? 
    \item[] Answer: \answerNA{} 
    \item[] Justification: This work does not involve crowdsourcing experiments or research with human subjects.

\item {\bf Institutional review board (IRB) approvals or equivalent for research with human subjects}
    \item[] Question: Does the paper describe potential risks incurred by study participants, whether such risks were disclosed to the subjects, and whether Institutional Review Board (IRB) approvals (or an equivalent approval/review based on the requirements of your country or institution) were obtained?
    \item[] Answer: \answerNA{} 
    \item[] Justification: This work does not involve human subjects or study participants, and therefore no IRB approval or participant risk disclosure was required.

\item {\bf Declaration of LLM usage}
    \item[] Question: Does the paper describe the usage of LLMs if it is an important, original, or non-standard component of the core methods in this research? Note that if the LLM is used only for writing, editing, or formatting purposes and does \emph{not} impact the core methodology, scientific rigor, or originality of the research, declaration is not required.
    \item[] Answer: \answerNA{} 
    \item[] Justification: LLMs were only used for writing, editing, and formatting assistance, and were not used as part of the core methodology, experiments, or scientific contributions of this research.

\end{enumerate}

\end{document}